\documentclass{article}

  \PassOptionsToPackage{numbers, compress}{natbib}
\usepackage[preprint]{neurips_2023}

\usepackage[utf8]{inputenc} %
\usepackage[T1]{fontenc}    %
\usepackage{url}            %
\usepackage{booktabs}       %
\usepackage{amsfonts}       %
\usepackage{nicefrac}       %
\usepackage{microtype}      %
\usepackage{xcolor}         %
\usepackage{color,xcolor}
\usepackage{epsfig}
\usepackage{graphicx}

\usepackage{adjustbox}
\usepackage{array}
\usepackage{booktabs}
\usepackage{colortbl}
\usepackage{float,wrapfig}
\usepackage{hhline}
\usepackage{multirow}
\usepackage{subcaption} %
\usepackage[font={small}]{caption}
\usepackage{natbib}

\usepackage{amsmath,amsfonts,amsthm,amssymb}
\usepackage{bm}
\usepackage{nicefrac}
\usepackage{microtype}
\usepackage{mathtools}

\usepackage{changepage}
\usepackage{extramarks}
\usepackage{fancyhdr}
\usepackage{lastpage}
\usepackage{setspace}
\usepackage{soul}
\usepackage{xspace}

\usepackage{url}

\usepackage{algorithmic}
\usepackage{algorithm}
\usepackage{todonotes} %
\usepackage{enumitem}
\usepackage{titlesec}
\usepackage{bbm}

\usepackage[pagebackref=true,breaklinks=true,colorlinks,citecolor=gray]{hyperref}

\newcommand{\sect}[1]{Section~\ref{#1}}

\newcommand{\fig}[1]{Figure~\ref{#1}}
\newcommand{\tbl}[1]{Table~\ref{#1}}

\newcommand{\myparagraph}[1]{\vspace{-8pt}\paragraph{#1}}

\hypersetup{colorlinks=true,linkcolor=red,citecolor=brown,urlcolor=blue}

\title{Improving Factuality and Reasoning in Language Models through Multiagent Debate}

\author{%
  Yilun Du \\
  MIT CSAIL\\
  yilundu@mit.edu\\
  \And
  Shuang Li \\
  MIT CSAIL\\
  lishuang@mit.edu \\
  \And
  Antonio Torralba \\
  MIT CSAIL \\
  torralba@mit.edu \\
  \And
  Joshua B. Tenenbaum \\
  MIT CSAIL, BCS, CBMM\\
  jbt@mit.edu \\
  \And
  Igor Mordatch \\
  Google Brain\\
  imordatch@google.com\\
}

\begin{document}

\maketitle

\begin{abstract}
    Large language models (LLMs) have demonstrated remarkable capabilities in language generation, understanding, and few-shot learning in recent years. An extensive body of work has explored how their performance may be further improved through the tools of prompting, ranging from verification, self-consistency, or intermediate scratchpads. 
    In this paper, we present a complementary approach to improve language responses 
    where multiple language model instances propose and debate their individual responses and reasoning processes over multiple rounds to arrive at a common final answer.
    Our findings indicate that this approach significantly enhances mathematical and strategic reasoning across a number of tasks. We also demonstrate that our approach improves the factual validity of generated content, reducing fallacious answers and hallucinations that contemporary models are prone to. Our approach may be directly applied to existing black-box models and uses identical procedure and prompts for all tasks we investigate.
    Overall, our findings suggest that such "society of minds" approach has the potential to significantly advance the capabilities of LLMs and pave the way for further breakthroughs in language generation and understanding. Project website at \url{https://composable-models.github.io/llm_debate/}.
    
\end{abstract}

\section{Introduction}

Large language models (LLMs) have demonstrated remarkable language generation, understanding, and few-shot learning capabilities in recent years. These methods are trained on a massive corpus of text on the internet, where the quality and accuracy of extracted natural language may not be ensured. Thus, current models may suffer from confidently hallucinating facts or making implausible jumps in chains of reasoning. An extensive body of recent work has focused on improving factual accuracy and reasoning in language models. These range from prompting models with few or zero-shot chain-of-thought demonstrations, use of verification, self-consistency, or intermediate scratchpads.

We note that these techniques are applied over a single model instance. Instead, we propose a complementary approach inspired by \emph{The Society of Mind} ~\citep{minsky1988society} and multi-agent settings, where multiple language model instances (or agents) individually propose and jointly debate their responses and reasoning processes to arrive at a single common answer. More specifically, given a query, multiple instances of a language model first generate individual candidate answers to a query. Then each individual model instance  reads and critiques the responses of all other models and uses this content to update its own answer. This step is then repeated over several rounds. This process induces models to construct answers that are consistent with both their internal critic as well as sensible in light of the responses of other agents. The resulting quorum of models can hold and maintain multiple chains of reasoning and possible answers simultaneously before proposing the final answer.

We find that our debate approach outperforms single model baselines such as zero-shot chain of thought ~\cite{kojima2022large} and reflection ~\cite{reflexion,madaan2023self} on a variety of six reasoning, factuality, and question-answering tasks. Using both multiple model agents and multiple rounds of debate are important to achieve the best performance. Given an initial query, we find that individual model instances propose a diverse range of answers despite being the same model class (although we also investigate the case of mixing different model types, such as chatGPT ~\cite{chatgpt2022} and Bard ~\cite{Pichai_2023}). After debating and examining the responses of other model instances, we find that the population almost always converges on a single and more accurate common answer. Debate results are also less likely to include false facts that models are internally uncertain of. This is because as the debate progresses, individual model instances tend to disagree on uncertain facts and omit them from the answer (Figure \ref{fig:validity_overview}). Lastly, we find that debate does not just act to amplify one correct answer in a model quorum - we find many cases where all the models initially make incorrect predictions, but then arrive at the correct answer as debate progresses (Figure \ref{fig:math_overview},\ref{fig:chatgpt_bard}).

We use the same methodology and prompt templates for all our tasks and require only black-box access to language model generations -- no model-internal information such as likelihoods or gradients is needed. This allows our method to be used with common public models serving interfaces. The method is also orthogonal to other model generation improvements such as retrieval or prompt engineering (in fact, we combine our debate method with zero-shot chain of thought). While the debate process is more costly, requiring multiple model instances and rounds, it arrives at significantly improved answers and may be used to generate additional model training data, effectively creating a model self-improvement loop.

To help evaluate the effect of our approach on factual accuracy, we introduce a new benchmark and dataset evaluating factual accuracy of famous computer scientist biographies. We find that contemporary language models have an especially high tendency to hallucinate factually incorrect biographies, often misrepresenting the relevant institutions and dates. Moreover, these facts often inconsistent across different language model instances. By asking models to come to a consensus across their answers, such inconsistent facts may be either removed or corrected.

In summary, our work contributes the following. First, we present a novel approach to improving factual correctness and reasoning accuracy in contemporary language models, leveraging a multi-agent debate process between models. Second, we introduce a new benchmark of factual correctness which contemporary language models struggle with. Finally, we evaluate the performance of our debate procedure in language generation, both in terms of the number of agents, the underlying rounds of debate, and the prompts that elicit such behavior across a set of six different reasoning and factual accuracy tasks.

\begin{figure}[t]
    \centering
    \includegraphics[width=1.0\linewidth]{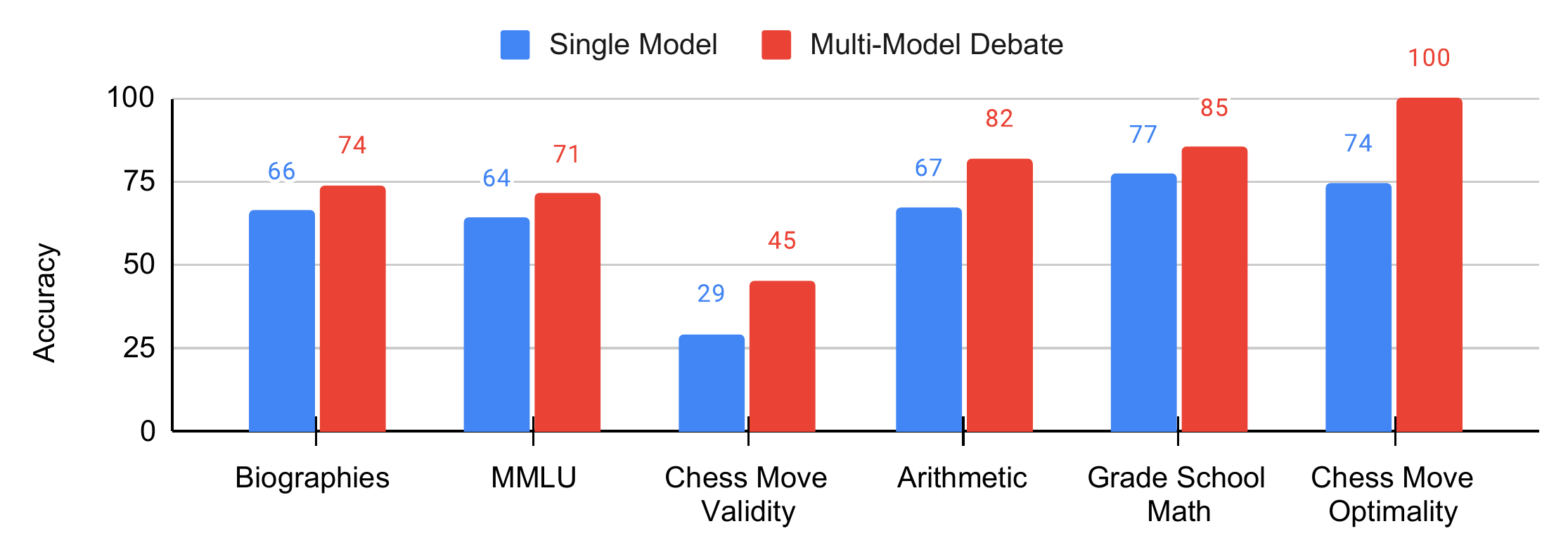}
    \vspace{-20pt}
    \caption{\textbf{Multiagent Debate Improves Reasoning and Factual Accuracy.} Accuracy of traditional inference and our multi-agent debate over six benchmarks (chess move optimality reported as a normalized score)}
    \label{fig:teaser}
    \vspace{-10pt}
\end{figure}

\vspace{-8pt}
\section{Language Generation through Multiagent Debate}
\vspace{-3pt}

\begin{figure}[t]
    \centering
    \includegraphics[width=\linewidth]{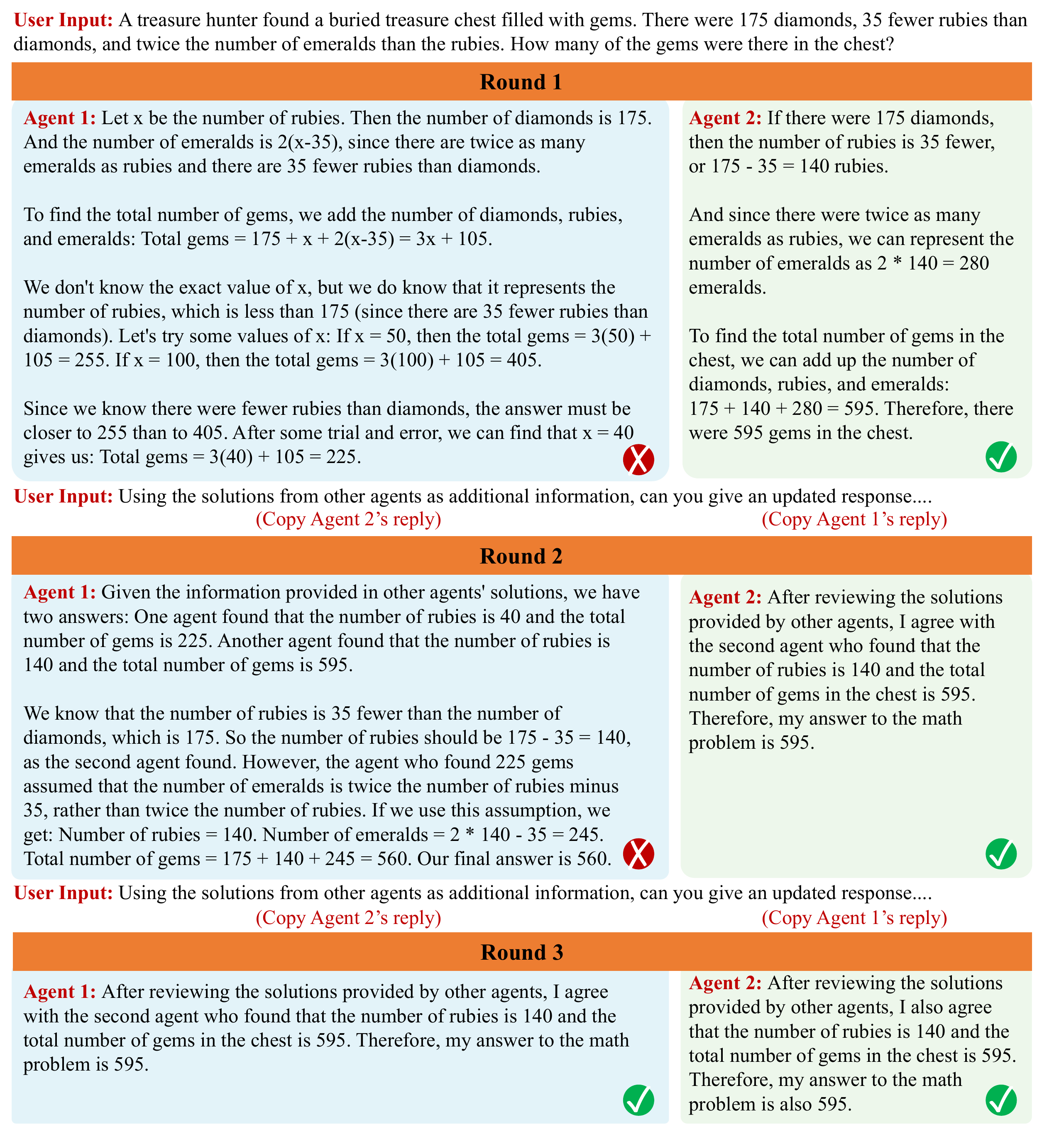}
    \caption{\textbf{Illustration of Debate.} Illustration of the debate procedure.}
    \label{fig:debate_overview}
    \vspace{-10pt}
\end{figure}

We present an approach to generate language responses through multiagent debate. We provide an overview of our approach in \sect{sect:multiagent}. We further discuss convergence to consensus in the debate process in \sect{sect:consensus}. The overall overview of our approach is shown in \fig{fig:debate_overview}.

\subsection{Multiagent Language Generation}
\label{sect:multiagent}
\vspace{-3pt}

Consider your work process when solving the following math question on an exam: {\it ``What is the area of a triangle with side lengths of 3, 4, 5?"}. In one thread of work, you may recognize that the triangle side-lengths directly correspond to a right triangle, and thus directly compute the area as $0.5 \times 3 \times 4 = 64$.  To make sure that you have the right answer, you may then try to solve the problem differently by estimating an angle $\theta$ in the triangle using the Law of Cosines, and then obtain the area by using the formula $0.5 \times 3 \times 4  \times \sin(\theta)$, arriving at another answer to the given exam problem.  

When these lines of work give the same answer, your confidence about the answer increases. In contrast, when these answers are different, individual lines of work may engage in a mental ``debate" procedure, where you closely cross-examine the reasoning and assumptions of each line of work and refine solutions  until a consistent answer.

Similarly, consider writing a biography of a historical figure. To ensure the factuality of the biography, you may consult multiple different sources on each fact. Facts that are consistent in each source increase your confidence about the fact. In contrast, facts that are inconsistent require careful cross-examination between sources to determine the final consistent data.

To mimic the above multi-threaded 
 reasoning process and multi-source factuality checking processes, we propose to generate answers subject to a multi-agent debate procedure between multiple instances of large language models. Given a question, multiple agents represented as copies of a large language model, generate answers to the question. Each response serves as a possible thought process or source of information which agents may re-examine to find consistent final answers.

After initial responses are generated from different agents, we initiate a round of debate between agents. Individual responses from other agents are concatenated and given as context to each agent, with each agent instructed to construct a new response based on such responses. Each language agent is thus responsible for both verifying the collection of responses given by other agents, and refining its own response based on other agents' responses. We iteratively repeat this debate procedure over multiple rounds for improved performance. 

Concretely, we first prompt each agent to independently solve the given problem or task. After each agent generates a response, we feed each agent a consensus prompt, illustrated in \fig{tbl:prompt}, where each agent is instructed to update their responses based on the responses of other agents. This resultant consensus prompt may then be repeatedly given, using the updated responses of each agent.
We illustrate an overview of this multiagent debate procedure in \fig{fig:debate_overview}.

Note that our proposed approach operates in an orthogonal manner to existing approaches to prompt language models. Given a question, we may apply additional techniques for prompting language models to further improve our debate procedure by eliciting additional more detailed responses from language models. We illustrate the synergy of our approach with existing approaches to prompting language models in \fig{fig:prompt_ablation} and directly apply zero-shot chain-of-thought reasoning in our evaluations.

\begin{table*}[t]
\small\setlength{\tabcolsep}{5.5pt}
\centering
\scalebox{0.9}{

\begin{tabular}{c|c}
     {\bf Debate Length} & {\bf Prompt} \\
      \midrule
     \multirow{2}{*}{Short}  & \emph{" These are the solutions to the problem from other agents: [other answers]} \\
      & \emph{Based off the opinion of other agents, can you give an updated response $\ldots$"} \\
      \midrule
      \multirow{2}{*}{Long}  & \emph{" These are the solutions to the problem from other agents: [other answers]}  \\
     & \emph{Using the opinion of other agents as additional advice, can you give an updated response $\ldots$"} \\
    \bottomrule
\end{tabular}
}
\captionof{figure}{\small {\bf Prompts to induce long and short form debate.} Responses of other agents to questions are are inserted in the middle of the prompt (indicated with \emph{[other answers]})}
\label{tbl:prompt}
\vspace{-10pt}
\end{table*}

\vspace{-5pt}
\subsection{Consensus in Debates}
\label{sect:consensus}

\begin{figure}[!t]
    \centering
    \includegraphics[width=\linewidth]{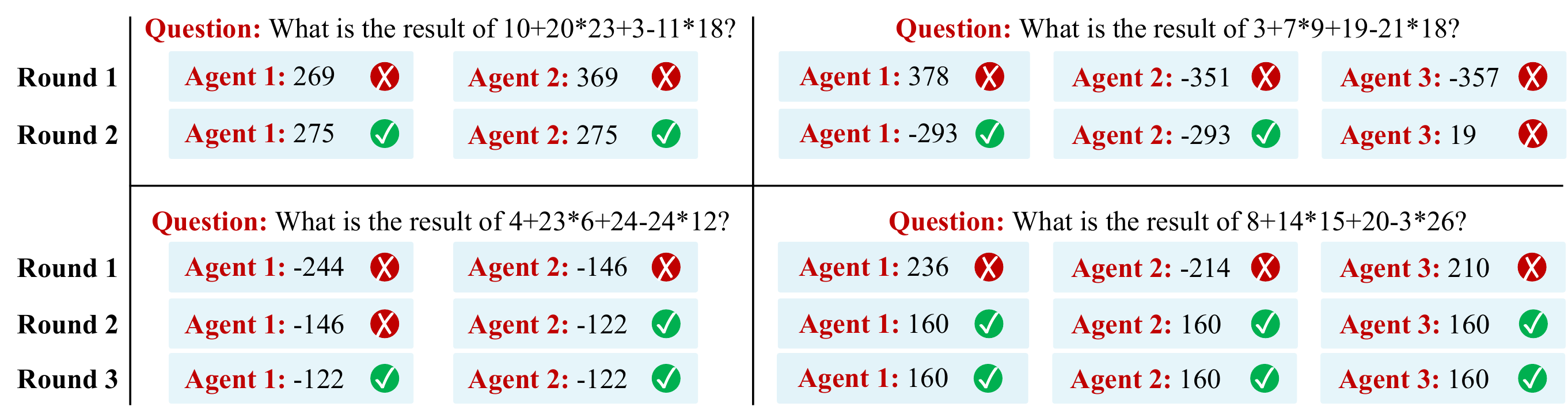}
    \caption{\textbf{Illustration of Solving Math.} Reasoning between agents is omitted.}
    \label{fig:math_overview}
    \vspace{-10pt}
\end{figure}

Given multiple rounds of debate, how can we ensure that a set of  language model agents will converge to a final consensus answer? In general, debate can be seen as a multi-agent game, where convergence is not guaranteed. Empirically, however, we find that language models are able to converge on a single shared answer after multiple rounds of debate (\fig{fig:math_overview}).

We found that we could control the duration of debates by how changing how much a language model trusts its own outputs over those generated by other models through different prompts. We illustrate two prompts below in \fig{tbl:prompt}, which we use to induce different debate durations between language models, and illustrate the effect of such prompts in \fig{fig:convergence}. In general, we found that prompts that encouraged models to be more ``stubborn’ based on their own solutions led to longer debates and better final solutions. Overall, we observed that language model agents were relatively "agreeable", perhaps as a result of instruction tuning or reinforcement learning based on human feedback ~\cite{ouyang2022training}.

\vspace{-8pt}
\section{Experiments}
\vspace{-3pt}

In our experiments, we evaluate our multiagent debate procedure and answer the following questions: \textbf{(1)} To what extent does multiagent debate improve reasoning? \textbf{(2)} To what extent does multiagent debate improve factual validity? \textbf{(3)}  What design choices enable multiagent debate to improve language generation performance?

\begin{figure}[t]
    \centering
    \includegraphics[width=0.9\linewidth]{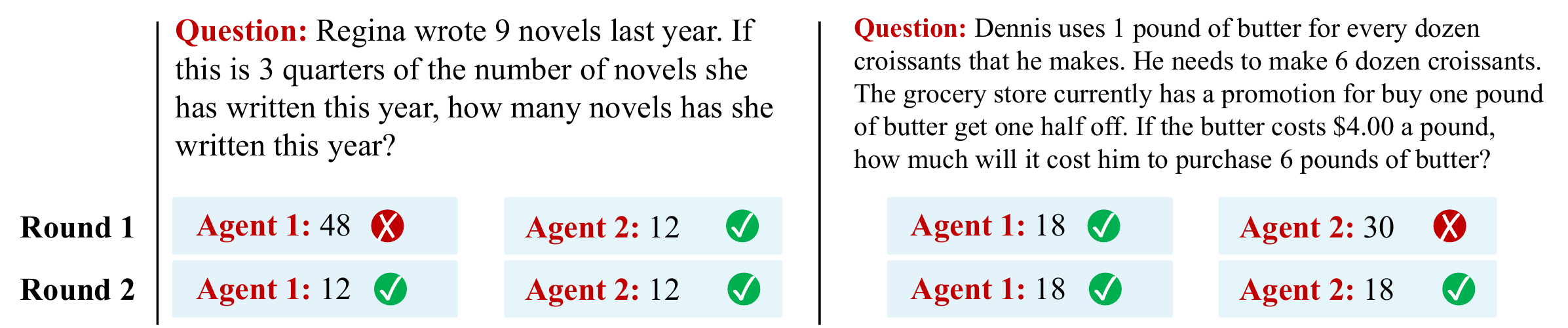}
    \caption{\textbf{Illustration of Solving Grade School Math.} Reasoning between agents omitted.}
    \label{fig:reasoning_overview}
    \vspace{-15pt}
\end{figure}

\begin{table*}[t]
\small\setlength{\tabcolsep}{5.5pt}
\centering
\begin{tabular}{lcccc}
      {\bf Model} & {\bf Arithmetic (\%) $\uparrow$} & {\bf Grade School Math (\%) $\uparrow$} & {\bf Chess ($\Delta$PS) $\uparrow$} \\
      \midrule
      Single Agent & 67.0 $\pm$ 4.7 & 77.0 $\pm$ 4.2 & 91.4  $\pm$ 10.6 \\
      Single Agent (Reflection) & 72.1 $\pm$ 4.5 & 75.0 $\pm$ 4.3 & 102.1 $\pm$ 11.9  \\
      Multi-Agent (Majority) & 69.0 $\pm$ 4.6   & 81.0 $\pm$ 3.9 & 102.2 $\pm$ 6.2 \\
      Multi-Agent (Debate) & \textbf{81.8 $\pm$ 2.3} & \textbf{85.0 $\pm$ 3.5}  & \textbf{122.9 $\pm$ 7.6} \\
    \bottomrule
\end{tabular}
\caption{\small \textbf{Multiagent Debate Improves Reasoning}  Multi-agent debate improves the reasoning abilities of language models. Multi-agent results in the table are run with 3 agents and two rounds of debate.}
\label{tbl:reasoning}
\vspace{-15pt}
\end{table*}

\vspace{-5pt}
\subsection{Improving Reasoning with Multiagent Debate}
\label{sect:reasoning}

We first evaluate the extent to which multiagent debate improves the underlying reasoning process in language models.

\myparagraph{Tasks.} We evaluate our approach on three reasoning tasks of increasing difficulty:
\vspace{-3pt}
\begin{itemize}[leftmargin=*, nosep]
    \item \texttt{Arithmetic.} We first evaluate the ability of models to correctly evaluate an arithmetic expression (containing addition, multiplication, and subtraction) consisting of six different two-digit numbers. For example: {\it What is the result of 12+15*21+0-3*27?} 
    \item \texttt{GSM8K.} Next, we consider harder mathematical reasoning tasks. Using the GSM8K dataset~\cite{verifier}, the models must correctly solve grade school mathematical reasoning tasks. 
    \item \texttt{Chess Move Prediction.} Finally, we consider the strategic reasoning of the ability of models, and ask models to predict the best next move in a game of chess, given the first 14 moves of a chess game between two chess grand-masters described in PGN notation~\cite{Fsmosca}. 
\end{itemize}
We report the accuracy of final answers in arithmetic and GSM8K tasks and report the pawn score (advantage) of predicted moves, as estimated by Stockfish in the Chess move prediction tasks. Additional details may be found in the Appendix.

\begin{wrapfigure}{r}{6.0cm}
    \vspace{-15pt}
    \centering
    \includegraphics[width=\linewidth]{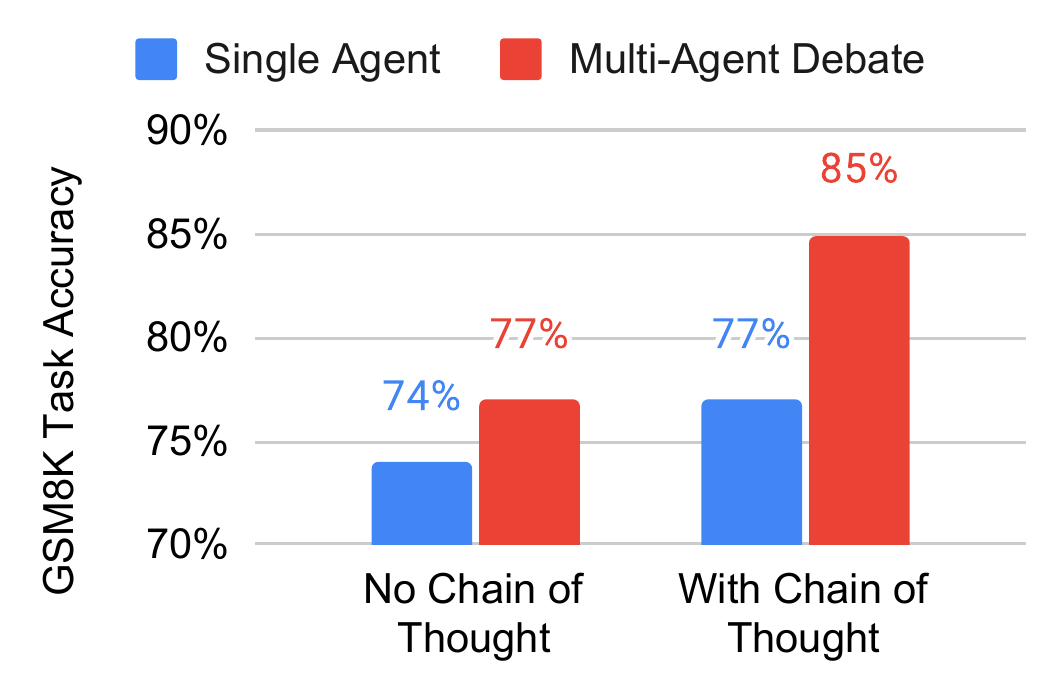}
    \vspace{-20pt}
    \caption{\textbf{Synergy with Other Methods.} Performance of debate increases with use of Chain of Thought prompting.}
    \vspace{-10pt}
    \label{fig:prompt_ablation}
\end{wrapfigure} 
\begin{figure}[t]
    \centering
    \includegraphics[width=0.9\linewidth]{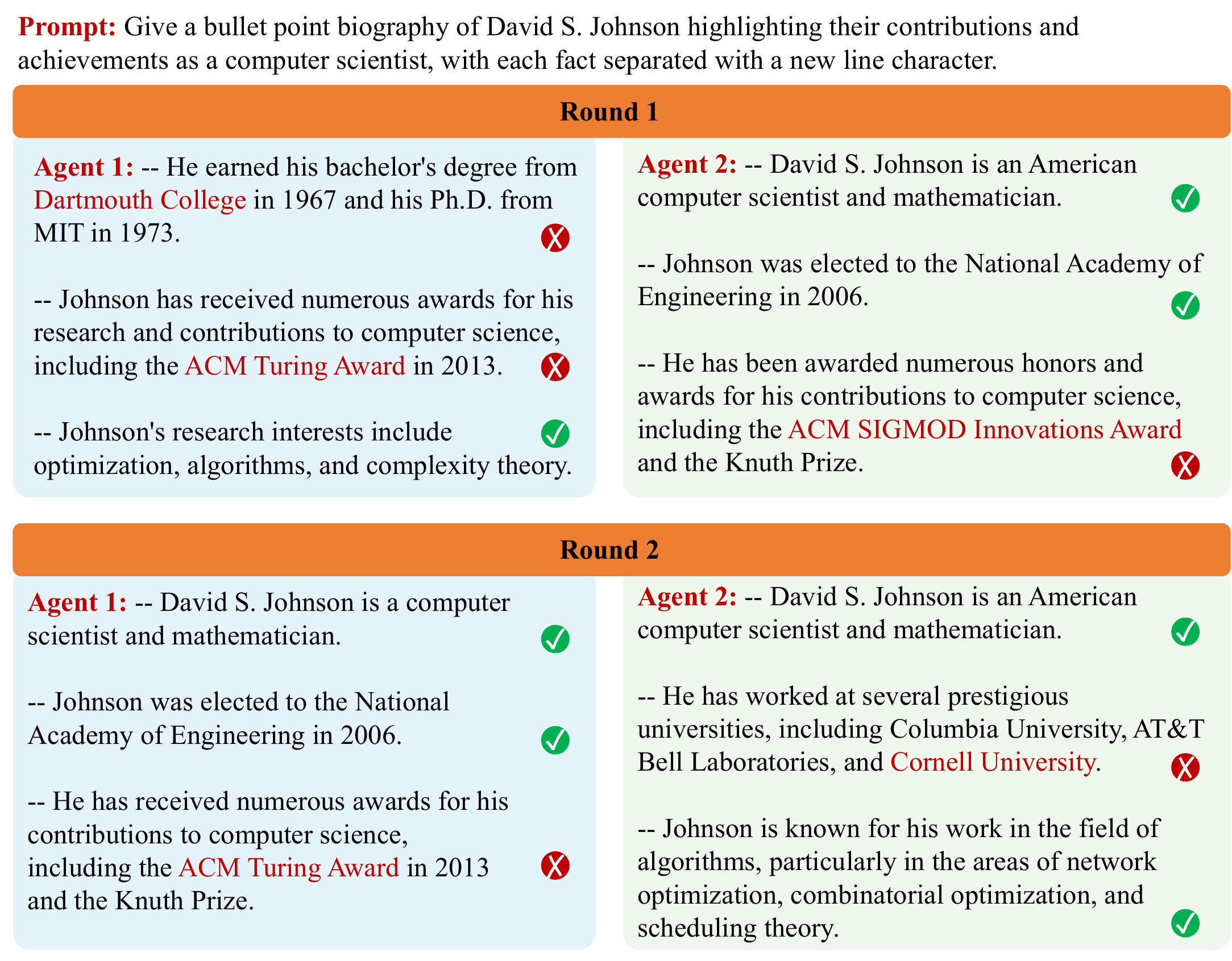}
    \caption{\textbf{Illustration of Generating Biographies.} Illustration of generating bullet  biographies of computer scientists. For brevity, only  the first 3 generated bullets are shown.}
    \label{fig:validity_overview}
    \vspace{-10pt}
\end{figure}
\begin{figure}[t]
    \centering
    \includegraphics[width=1\linewidth]{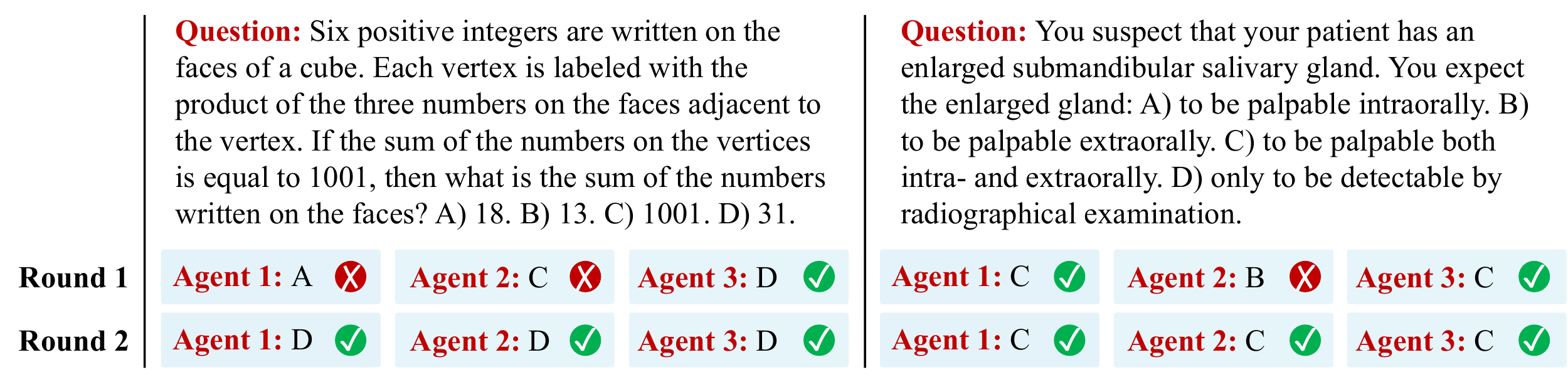}
    \caption{\textbf{Illustration of MMLU.} Illustration of debate when answering factual tasks. Reasoning omitted.}
    \label{fig:mmlu_result}
    \vspace{-10pt}
\end{figure}

\begin{figure}[t]
    \centering
    \includegraphics[width=1\linewidth]{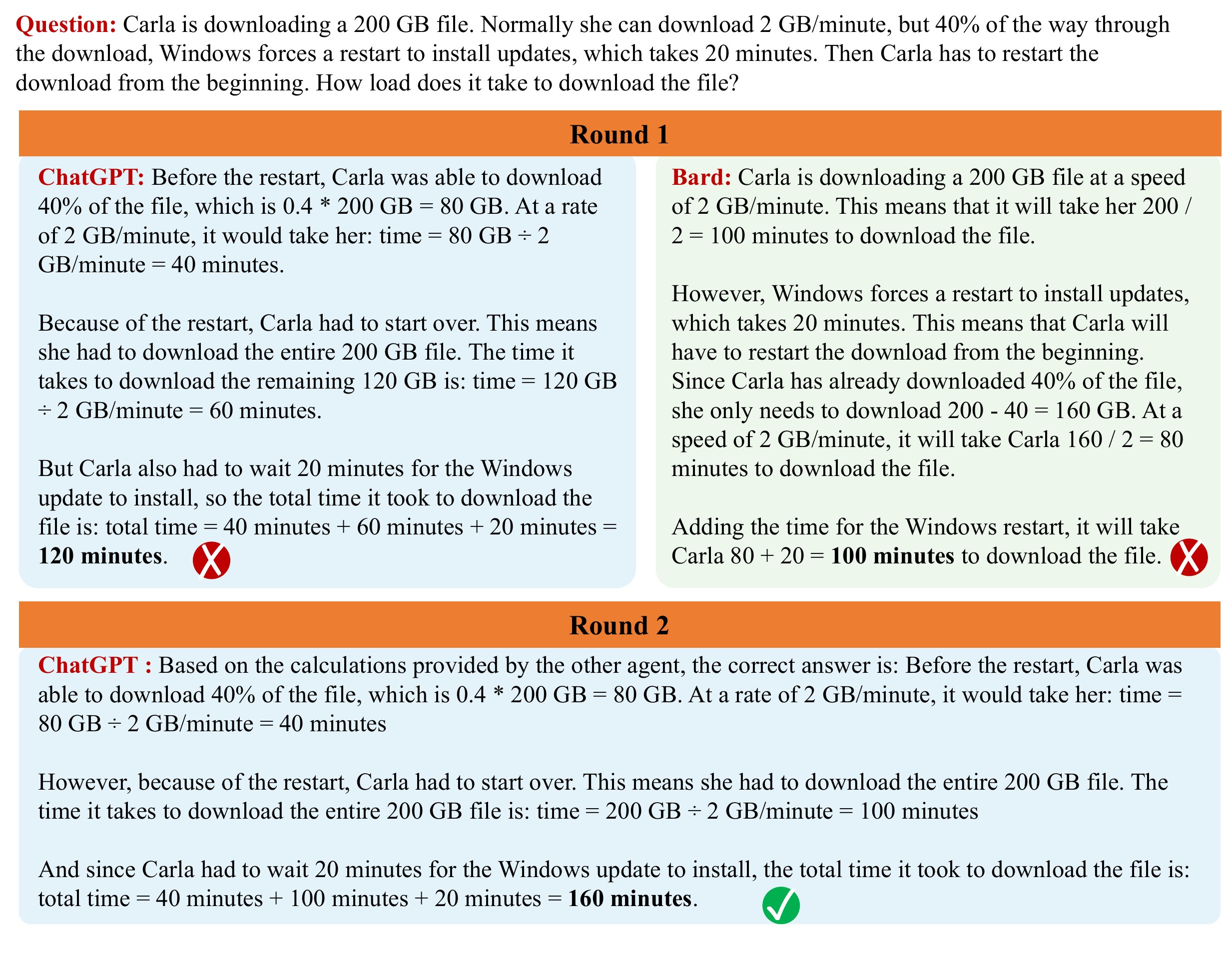}
    \caption{\textbf{Debate Between chatGPT and Bard} Illustration of debate between different models.}
    \label{fig:chatgpt_bard}
    \vspace{-15pt}
\end{figure}

\myparagraph{Baselines.} We compare our approach to three alternative approaches to generate responses for reasoning problems. First, we ask agents to directly generate responses (single agent). Next, we consider asking language models to generate and then "self-reflect" on the responses generated~\cite{reflexion, madaan2023self}. Finally, we consider generating responses using multiple agents and performing majority voting~\cite{alphacode,verifier}. As the focus of our experiments is to verify the effectiveness of multiagent agent debate, we run both baselines and our approach, using the identical starting prompt and language model across all evaluations. We evaluate models in a zero-shot setting, with prompts found in the Appendix of the paper. We use chatGPT-based language model ~\cite{chatgpt2022} in all our experiments except those in \fig{fig:chatgpt_bard} where we compare multiple language models.

Due to computational expense, we evaluate our approach across benchmarks mainly using three agents with two rounds of debates, although we found further gains with both more agents and rounds of debate (\fig{fig:round_agents}). Additional evaluation details are found in the Appendix.

\myparagraph{Quantitative Results.} In \tbl{tbl:reasoning}, we report the results of each approach on arithmetic, grade school math, and chess reasoning task. In each task, we observe that utilizing multiple different agents to generate solutions improves performance over using a single language model agent to generate a solution. Simultaneously, we also see that reflection, where a language model is asked to critique its early generation, generally gives a modest boost in performance. Multiagent debate, which may be seen as a combination of both reflection and multiagent generation, gives a substantial boost in reasoning across each of the tasks.

\myparagraph{Qualitative Results.} In Figure ~\ref{fig:math_overview}, \ref{fig:reasoning_overview}, we provide qualitative illustrations of the debate procedure between models. Interestingly, we find cases in which all models initially give an incorrect response, yet the result of debate still obtains the correct answer as agents critique each others' reasoning. Thus, the purpose of our debate isn't just to amplify a correct answer -- all models can initially be wrong but arrive at the correct answer through the debate process.

\myparagraph{Compatibility with other reasoning methods.} Our multiagent generation procedure operates orthogonally approach to other prompting methods which focus on single-agent generation. In \fig{fig:prompt_ablation}, we illustrate the performance of multi-agent debate with and without zero-shot chain-of-thought prompting~\cite{kojima2022large} on GSM8K. In both settings, multiagent generation is beneficial.

\begin{table*}[t]
\small\setlength{\tabcolsep}{5.5pt}
\centering
\begin{tabular}{lcccccc}
      {\bf Model} & {\bf Biographies} & {\bf MMLU} & {\bf Chess Move Validity} \\
      \midrule
      Single Agent & 66.0 $\pm$ 2.2 & 63.9 $\pm$ 4.8 & 29.3 $\pm$ 2.6\\
      Single Agent (Reflection) & 68.3 $\pm$ 2.9 & 57.7 $\pm$ 5.0 & 38.8 $\pm$ 2.9 \\
      Multi-Agent (Debate) & \textbf{73.8 $\pm$ 2.3} & \textbf{71.1 $\pm$ 4.6} & \textbf{45.2 $\pm$ 2.9}   \\
    \bottomrule
\end{tabular}
\caption{\small \textbf{Multiagent Debate Improves Factual Accuracy}  Multi-agent debate improves the factual accuracy.}
\label{tbl:accuracy}
\vspace{-5pt}
\end{table*}

\begin{figure}[t]
    \centering
    \includegraphics[width=1.0\linewidth]{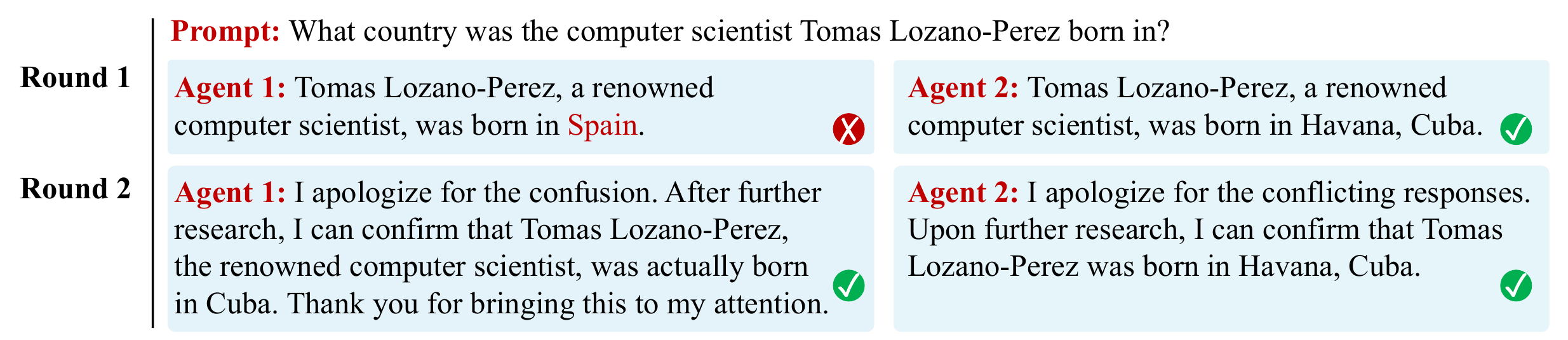}
    \caption{\textbf{Expressing Uncertainty with Multiple Answers.} For facts that a language model is uncertain about, different language agents generate different facts. Debate causes agents to converge to one fact that is more accurate, but not necessarily always factually correct.}
    \label{fig:uncertainty_generation}
    \vspace{-10pt}
\end{figure}
\begin{figure}[!t]
    \centering
    \includegraphics[width=\linewidth]{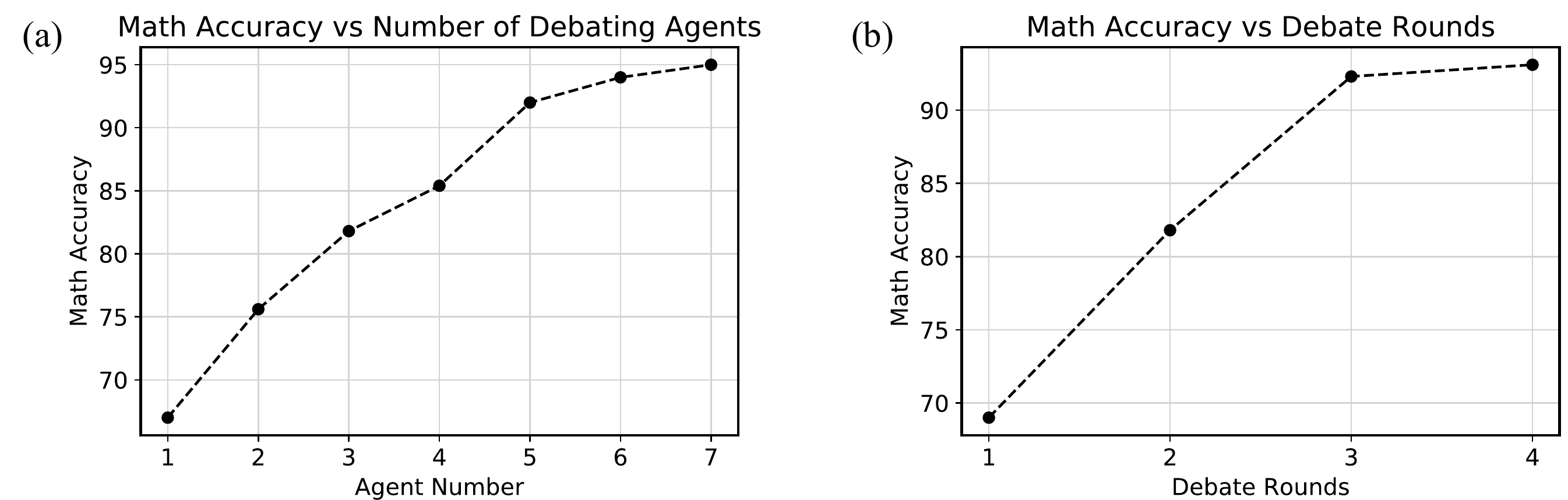}
    \caption{\textbf{(a) Performance with Increased Agents.} Performance improves as the number of underlying agents involved in debate increases. \textbf{(b) Performance with Increased Rounds.} Performance rises as the number of rounds of underlying debate increases.}
    \label{fig:round_agents}
\vspace{-15pt}
\end{figure}

\vspace{-3pt}
\subsection{Extracting Factual Information from Multiagent Debate}

We next evaluate the extent to which multiagent debate improves the underlying factuality in language models.

\myparagraph{Tasks.} We evaluate the factuality of language models in three different settings:

\begin{itemize}[leftmargin=*,nosep]
    \item \texttt{Biographies.} To evaluate the factuality of language models, we introduce a novel task of accurately generating historical biographies of people. In preliminary testing, we found that existing language models had a tendency to hallucinate many facts on this task. We constructed ground truth bullet point biographies of 524 well-known computer scientists. We then asked language models to generate bullet point biographies for each person, and evaluated the accuracy at which each ground truth bullet point agreed with generated bullets. We report additional evaluation details in the Appendix.
    \item \texttt{MMLU.} Next, we assess the factuality of language models in responding to different factual knowledge questions typically learned and assessed in different exams. We utilize the existing MMLU dataset~\cite{hendrycks2020measuring}  to benchmark the accuracy of responses. 
    \item \texttt{Chess Move Validity.} Lastly, we study the hallucinations in language models when planning under to the given rules of an existing environment or game. Specifically, we measure the validity of possible moves in a game of Chess given by BIG-Bench Chess-State Tracking Benchmark ~\cite{srivastava2022beyond} task of chess-move prediction. In this task, an agent is given a set of next moves, and must make a valid next move of a piece on a board.
\end{itemize}

\myparagraph{Baselines.} We use the same baselines as in \sect{sect:reasoning}. The multiagent (majority) is not directly applicable in this setting as individual responses are not easily comparable, and so we omit baseline comparison with the majority voting in this setting.

\myparagraph{Results.} We analyze the performance of each method in \tbl{tbl:accuracy}. We found that approaches based on reflection led to poor performance in the factuality setting. In contrast, debate gives the best performance in this setting also, and significantly outperforms each baseline. We illustrate a debate between agents on the biography task in \fig{fig:validity_overview} and on MMLU in \fig{fig:mmlu_result}. We found that multiagent debate improved and settled on bullets that were more consistent across agents. 

We found that different language agents tended to give different answers when the underlying language model was uncertain about the question. However, directly asking each agent about their confidence ~\cite{kadavath2022language} of the answer led to high confidence assessments on each answer. However, when these different language agents were asked to communicate with each other, each agent would quickly change their opinion to a consensus answer which was more accurate.  We illustrate this in \fig{fig:uncertainty_generation}. Interestingly, we found that on facts that the language model was confident in (i.e. many instances of the same model all gave the same answer), it was very difficult to convince an agent to change their opinion, suggesting that ``ease of persuasion'' may be a method to assess factual confidence.

\vspace{-5pt}
\subsection{Analysis: Understanding Multiagent Debate}

Finally, we analyze how multiagent debate improves the underlying language generation procedure in language models.

\myparagraph{Number of Agents.} First, we analyze the impact of agents number in debate. In \fig{fig:round_agents}(a), we increase the number of agents used in debate, while fixing the debate length to be two. On arithmetic, performance monotonically increases with the increased number of agents. For larger number of agents, we first summarize all agent responses with chatGPT instead of directly concatenating responses due to context length error.

\myparagraph{Rounds of Debate} Next, we analyze the impact of the number of rounds of debate in multiagent debate. In \fig{fig:round_agents}(b), we increase the debate length between agents, while fixing the number of agents to three. We find that on the arithmetic task, the performance also monotonically increases with debate length. However, we found that additional debate rounds above four led to a similar final performance to 4 rounds of debate.

\begin{wrapfigure}{r}{6.0cm}
    \vspace{-20pt}
    \centering
    \includegraphics[width=\linewidth]{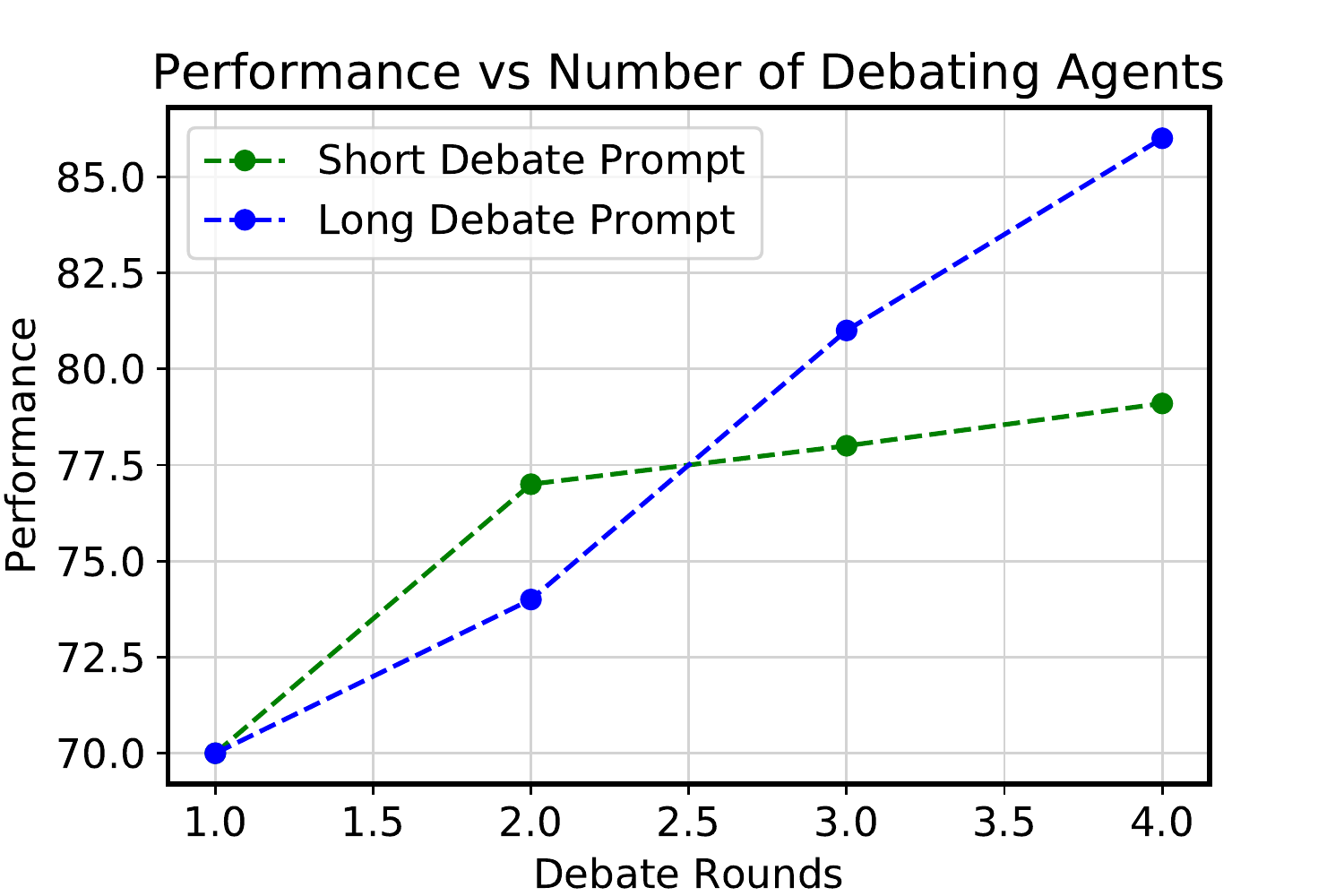}
    \vspace{-15pt}
    \caption{\textbf{Performance vs Debate Length.} Prompts which induce longer debate improve performance.}
    \vspace{-15pt}
    \label{fig:convergence}
\end{wrapfigure}

\myparagraph{Effect of Debate Length on Accuracy} As discussed in \sect{sect:consensus}, the underlying convergence time in the debate between agents can be controlled by the extent to which agents are encouraged to maintain their opinions. 
In \fig{fig:convergence}, we consider the effect of short and long-form prompts discussed in \fig{tbl:prompt}. We find that debates using longer prompts lead to slower convergence to correct answers, but also lead to a better final consensus on the correct answer. We provide an analysis of consensus between agents in \fig{fig:consensus}.

\myparagraph{Using Different Initialization Prompts} In our experiments we use the same prompts for all agents. We also consider the effect of using different questions, where we first instruct each language model to behave like a different persona (professor, doctor, mathematician) on the MMLU dataset. We found that improved performance on MMLU from 71.1 to 74.2 with different agents, suggesting further gains can be obtained with different initialization prompts.

\begin{wrapfigure}{r}{6.0cm}
    \vspace{-20pt}
    \centering
    \includegraphics[width=\linewidth]{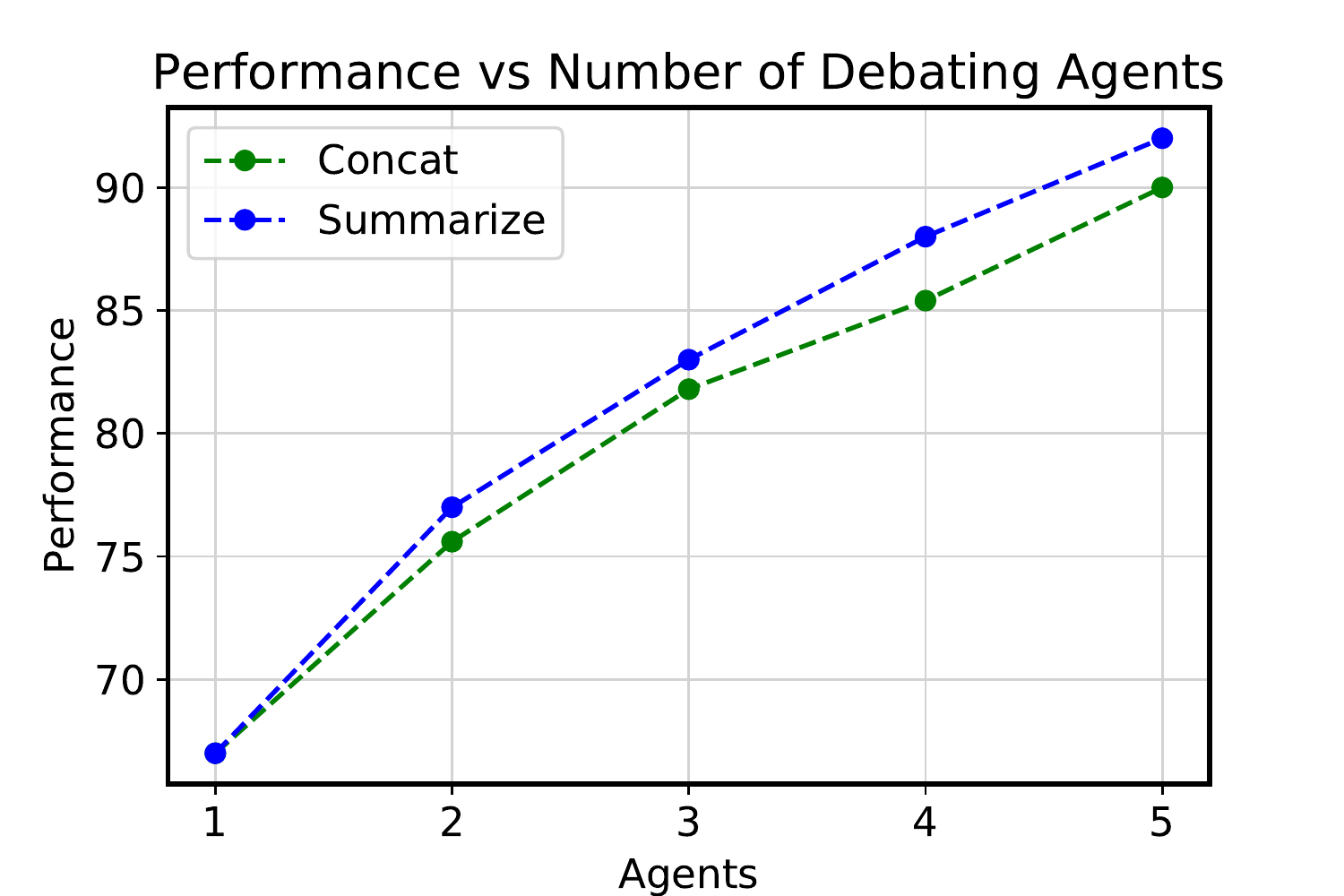}
    \vspace{-15pt}
    \caption{\textbf{Effect of Summarization.} When there are many agents in a debate, responses from other agents may be first summarized and then given as context, reducing context length. This operation improves performance.}
    \vspace{-15pt}
    \label{fig:summarize}
\end{wrapfigure}

\myparagraph{Summarization.} While in the majority of experiments in the paper we directly concatenate the responses of other agents as context for an agent to generate a new response, this is expensive when the number of agents involved in debate gets large. We may alternatively first summarize the responses from all other agents into a single response that we provide to agent at each round for more efficient debate. We apply this strategy in \fig{fig:round_agents} to enable the use of five or more agents in debate. In \fig{fig:summarize}, we analyze the effect compared to directly concatenating the responses of other agents. We find this improves the performance of debate, suggesting that summarization is another tool that can further improve multiagent debate.

\myparagraph{Utilizing Different Language Models} Our existing debate results are reported using multiple instances of a chatGPT language model. We further assess the impact of using two different language models, where we ask chatGPT and Bard ~\cite{Pichai_2023} language models to debate with each other on a set of 20 GSM8K math problems.  In this set, we find that multi-agent debate improves the performance of both agents, with Bard solving 11 problems, chatGPT solving 14 problems, and joint multi-agent debate solving 17 problems. We qualitatively illustrate a debate between agents in \fig{fig:chatgpt_bard}. While both agents initially provide incorrect answers to the problem, chatGPT is able to utilize the incorrect response given by Bard to generate the final correct answer.

\vspace{-8pt}
\section{Related Work}

\myparagraph{Reasoning and Factuality in Language Models.} A wide range of work has explored how to enable reasoning and factuality in language models. To improve reasoning, approaches have relied on prompting techniques such as scratchpads ~\cite{nye2021show},  verification ~\cite{verifier}, chain-of-thought demonstrations ~\cite{wei2022chain,kojima2022large,reynolds2021prompt}, and intermediate self-reflection ~\cite{reflexion, madaan2023self} and finetuning~\cite{lewkowycz2022solving, rajani2019explain,zelikman2022star}. To improve factuality, approaches have relied on training techniques such as RLHF ~\citep{ziegler2019rlhf,liu2022instruction,christiano2017rlhf}, pruning truthful datasets~\citep{lee2022factuality}, external knowledge retrieval~\citep{guu2020realm} and training-free methods based off likelihood estimation ~\citep{kadavath2022language}.

Our work provides an alternative way to obtain reasoning and factuality in language models using multiagent debates, which only requires black-box access to a language generator. Prior work also has explored how to take the majority vote across different models ~\cite{alphacode,verifier,wang2022self,thoppilan2022lamda} while in this work, we use the power of a language model to combine different answers. Most similar to our work, \citet{irving2018ai} also proposes a debate procedure to verify the accuracy and safety of powerful AI agents. In contrast to our approach, in their work, agents are asked to alternatively provide proof of a input, and humans are tasked with assessing these debates and determining safety.

\myparagraph{Compositional Generation.} Our work is also related to existing works that focus on text generation by combining different models ~\cite{du2020compositional, liu2022compositional,zeng2022socratic,alayrac2022flamingo,du2023reduce}. Most similar to our work, \cite{li2022composing,zeng2022socratic} propose to combine multiple different large pretrained models together for multimodal reasoning.  In contrast, in our work, we aim to use communication between different language models to enable more effective reasoning and factuality in language models. 
\vspace{-8pt}
\section{Limitations and Discussion}
\vspace{-3pt}

In this paper, we present an orthogonal approach to improve the performance of language models using multi-agent debate. We find that the approach is simple and effective across a wide set of different reasoning and validity language modeling tasks.

\myparagraph{Limitations.} In comparison to other prompting techniques, our multiagent debate procedure is more computationally expensive, as it requires both multiple language generations, and an underlying debate procedure. However, we believe that this approach may be seen as a method to generate additional data that may be distilled back to self-improve the original base model.

Further, we observed that as debates became longer in duration, current language models sometimes struggled to fully process the entire debate input, and typically only focused on the most recent generations. We believe that this performance will be alleviated with longer-context and improved language models or by summarizing early portions of the debate.

Finally, we found that while debates typically converged into single final answers, these answers were not necessarily correct. Despite answers being incorrect, language models would confidently affirm that their answer is correct and consistent with all other agent responses. We believe this result is in part due to the fact that LMs do not correctly express their uncertainty when generating responses, and believe that other orthogonal approaches to improve this performance would improve the results of multiagent debate.

\bibliographystyle{abbrvnat}
\bibliography{main}

\newpage
\appendix

\section{Appendix}

In this appendix, we provide additional analysis and visualizations of the debates used in the main paper in \sect{sect:additional_results}. We further provide detailed experimental details on each dataset in \sect{sect:evaluation}.

\subsection{Additional Results}
\label{sect:additional_results}

\begin{wrapfigure}{r}{6.0cm}
    \vspace{-20pt}
    \centering
    \includegraphics[width=\linewidth]{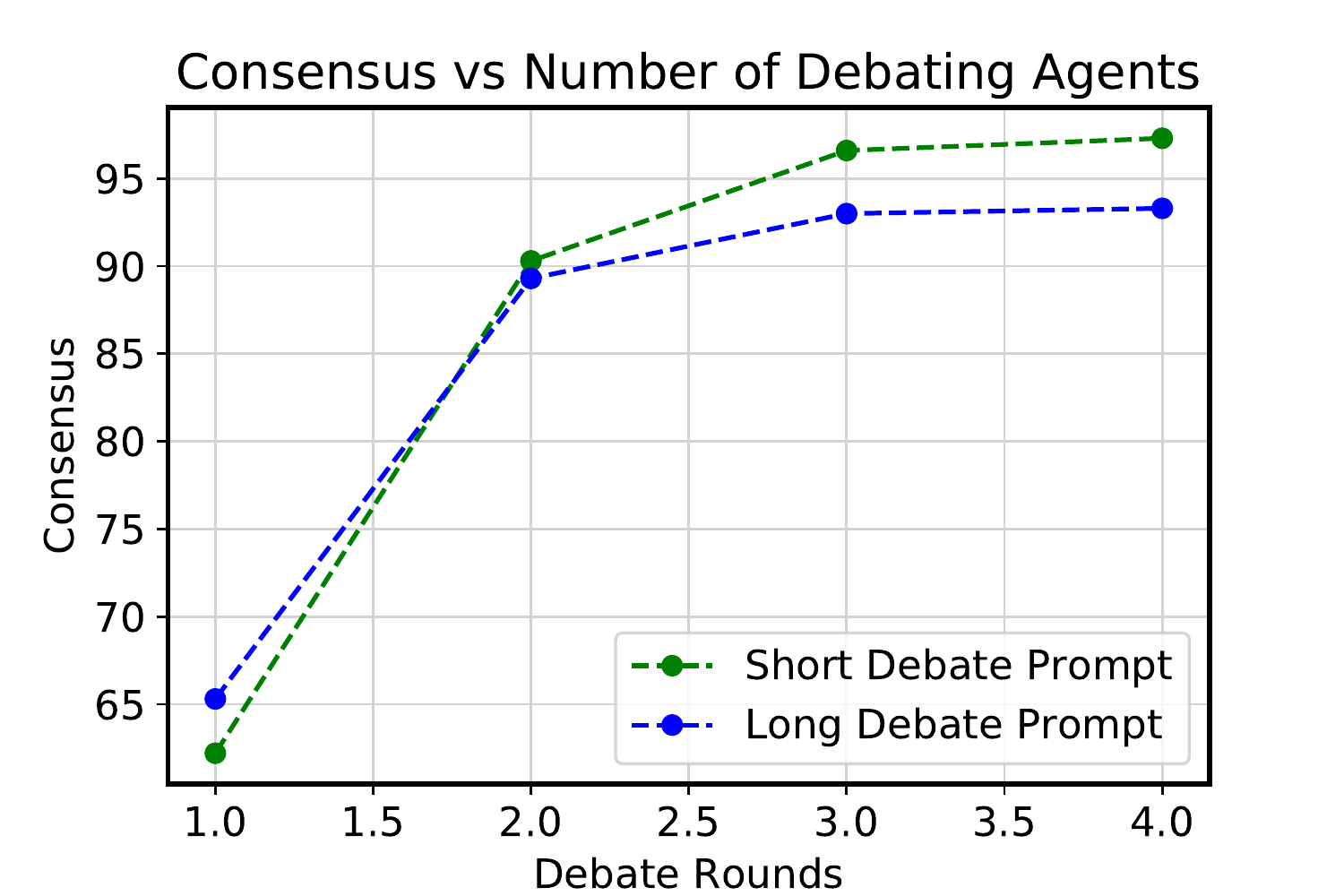}
    \vspace{-15pt}
    \caption{\textbf{Effect of Prompts on Consensus.} Using a short debate prompt induces faster consensus between agents}
    \vspace{-15pt}
    \label{fig:consensus}
\end{wrapfigure}

\vspace{3pt}
\myparagraph{Consensus Between Agents.} In \fig{fig:consensus}, we illustrate the consensus between agents using either short or long consensus prompts discussed in \fig{tbl:prompt}. The use of debate prompts that encourage agents to adapt more to the opinions of other agents improves consensus.

\begin{figure}[t]
    \centering
    \includegraphics[width=\linewidth]{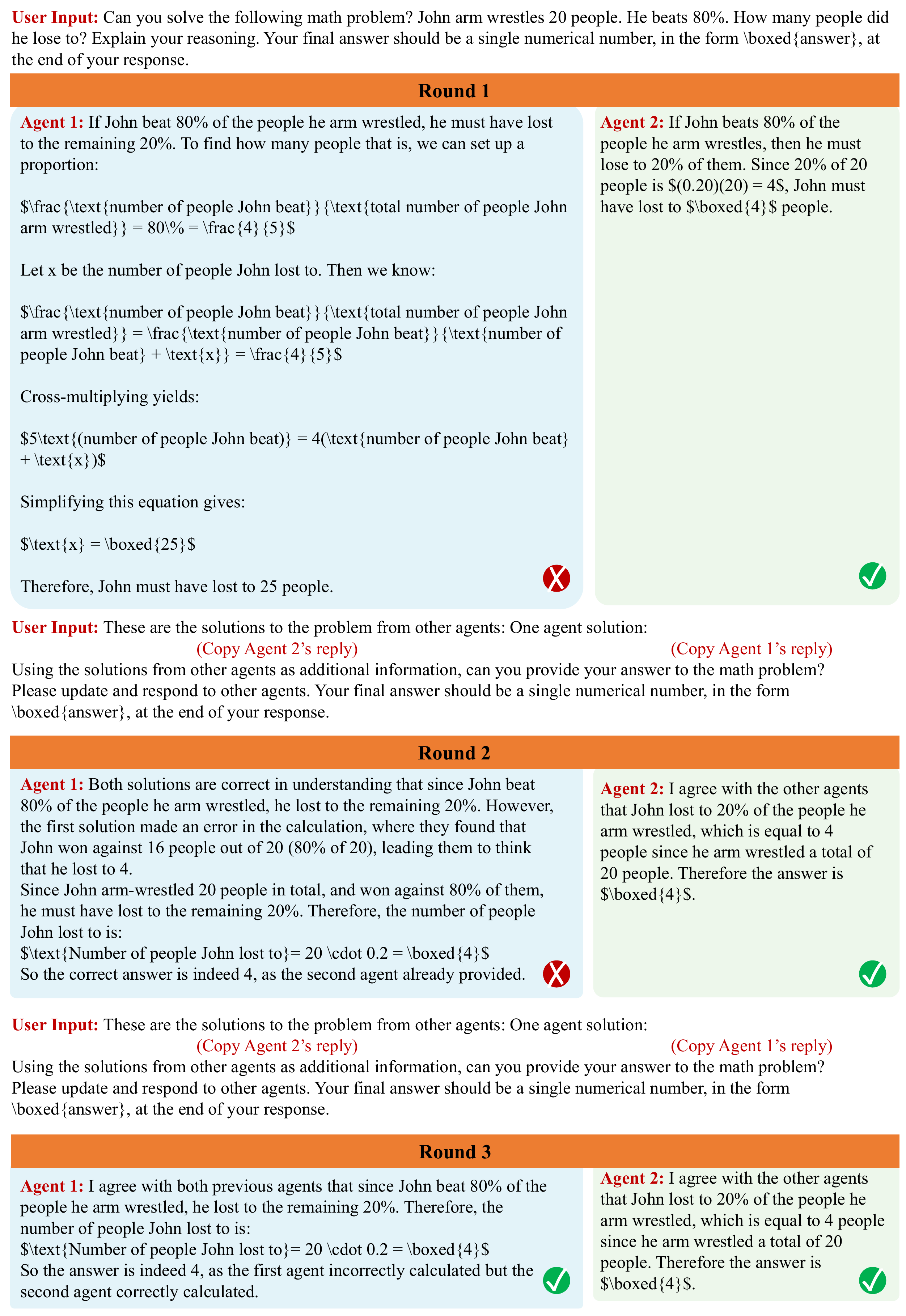}
    \caption{\textbf{Example of a correct GSM8K Debate.}}
    \label{fig:gsm-1}
    \vspace{-10pt}
\end{figure}

\begin{figure}[t]
    \centering
    \includegraphics[width=\linewidth]{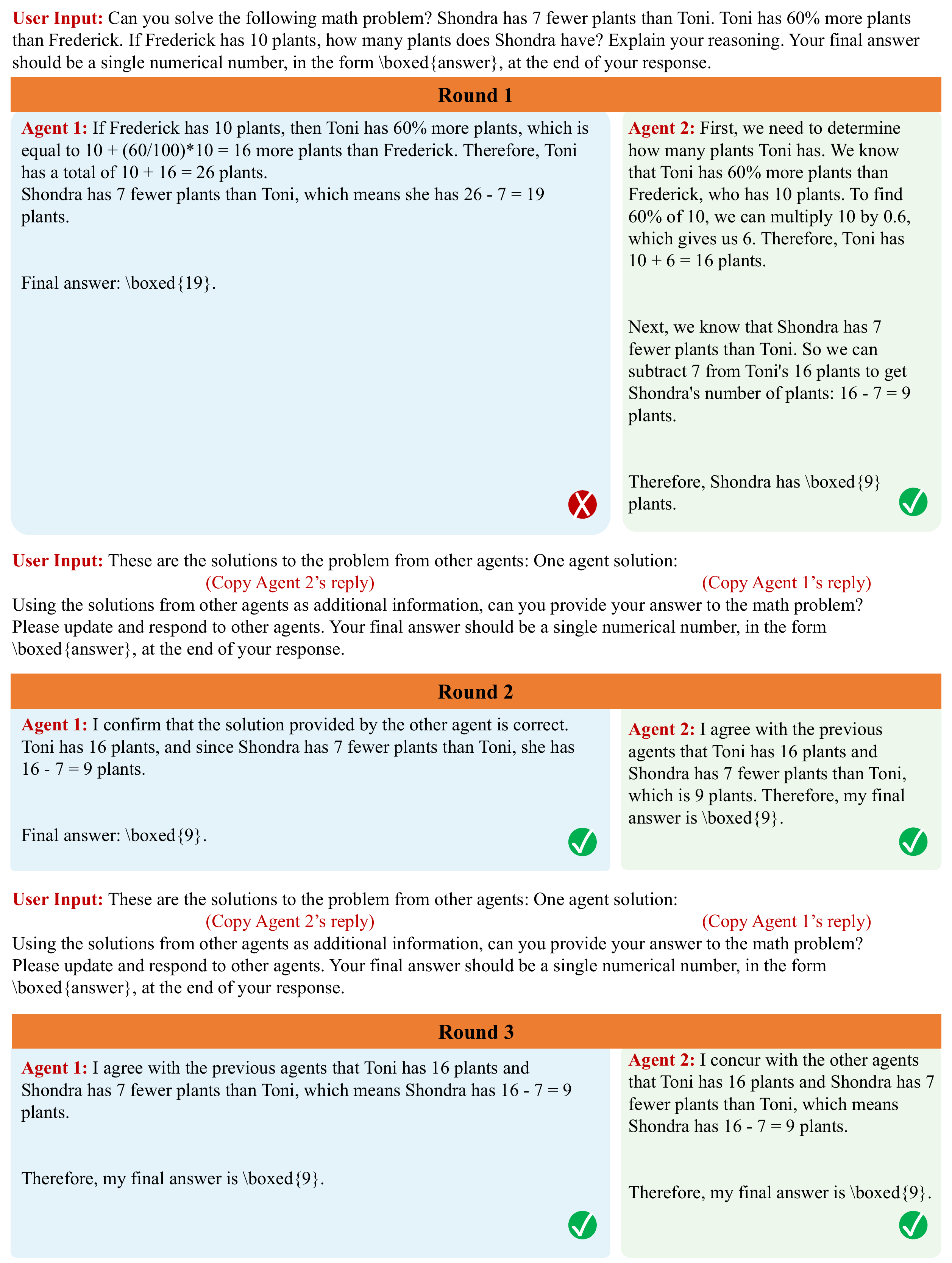}
    \caption{\textbf{Example of Correct GSM8K Debate.}}
    \label{fig:gsm-2}
    \vspace{-10pt}
\end{figure}

\begin{figure}[t]
    \centering
    \includegraphics[width=\linewidth]{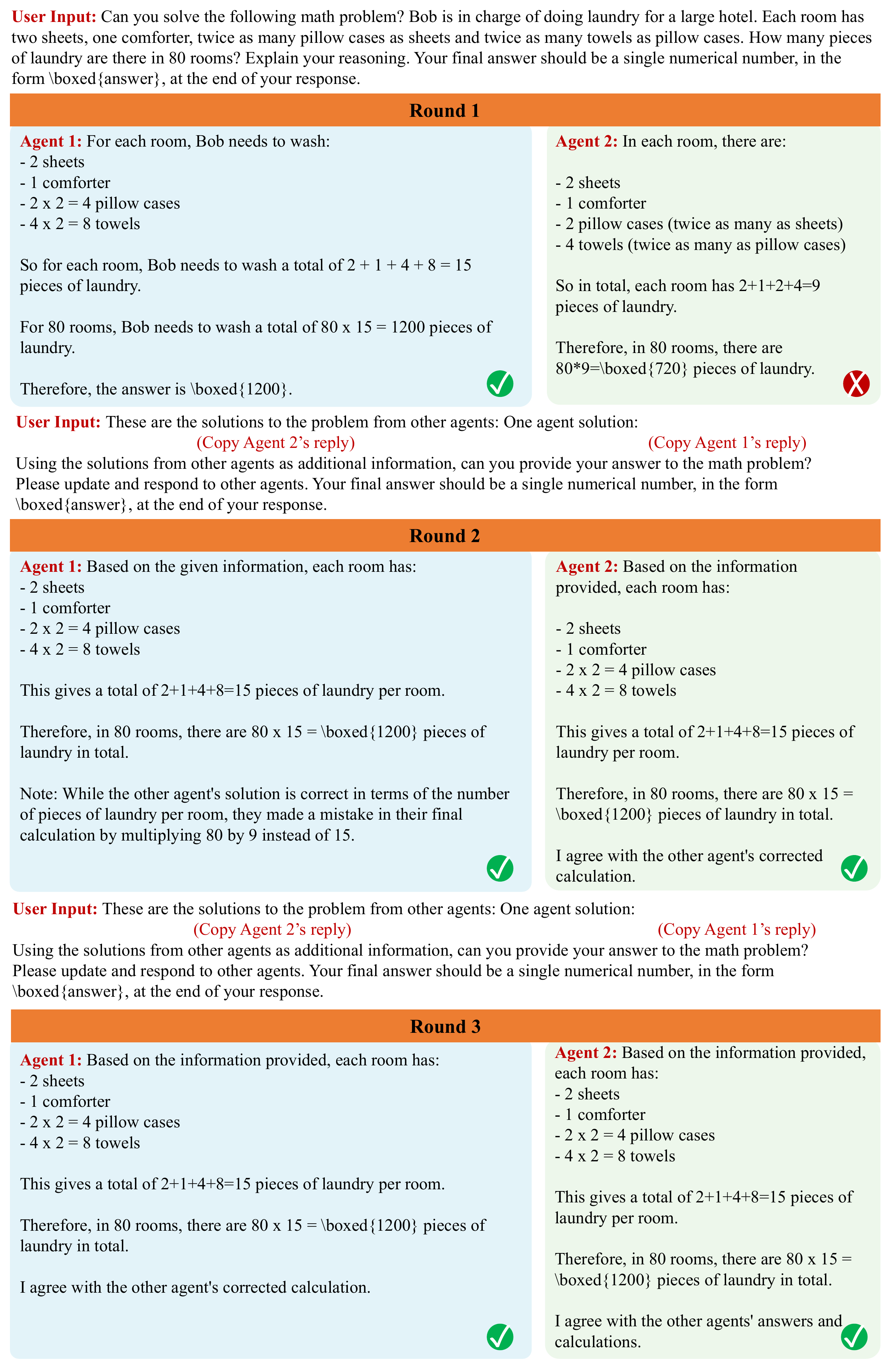}
    \caption{\textbf{Example of Correct GSM8K Debate.}}
    \label{fig:gsm-3}
    \vspace{-10pt}
\end{figure}

\begin{figure}[t]
    \centering
    \includegraphics[width=\linewidth]{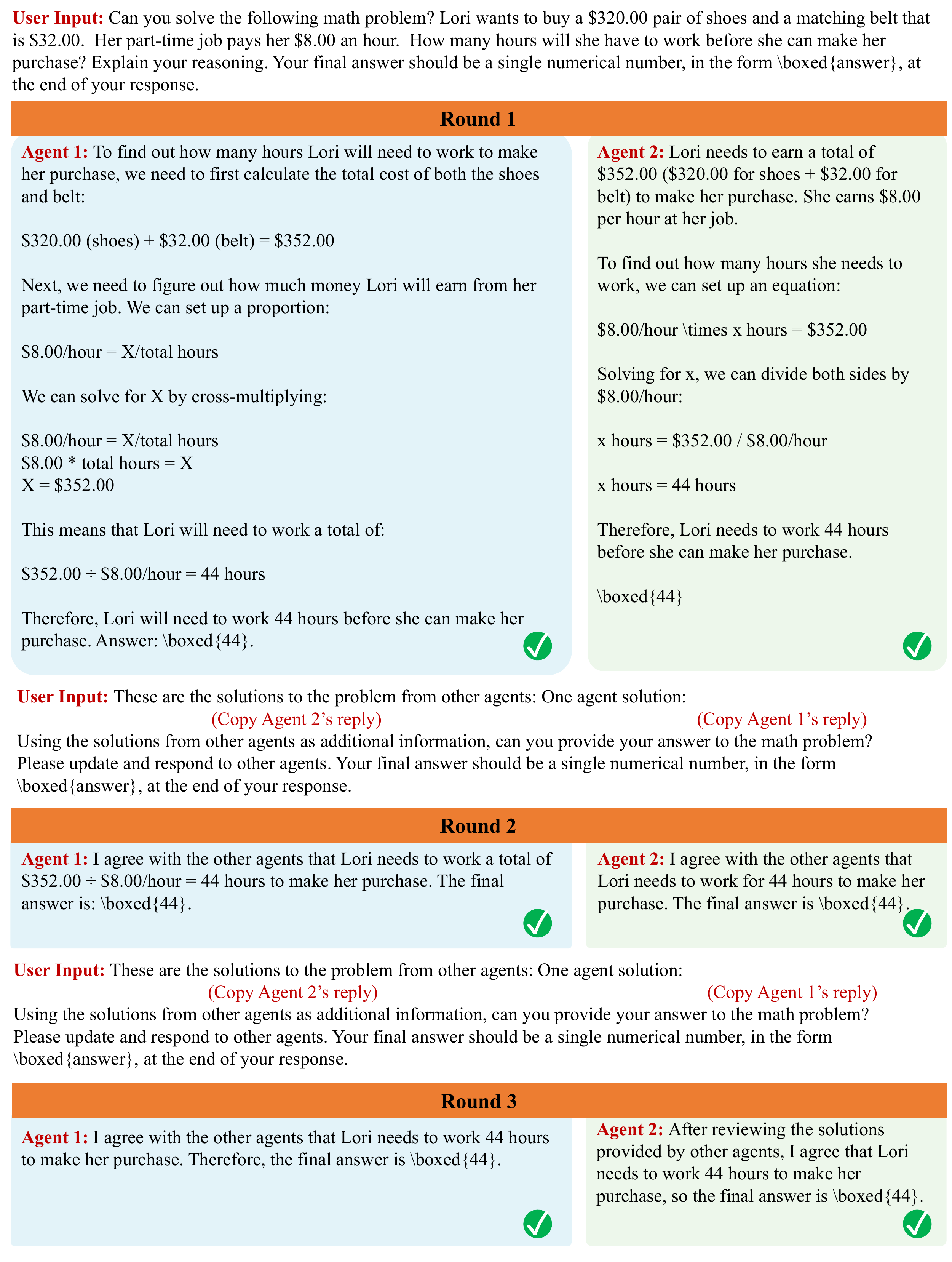}
    \caption{\textbf{Example of Correct GSM8K Debate.}}
    \label{fig:gsm-4}
    \vspace{-10pt}
\end{figure}

\begin{figure}[t]
    \centering
    \includegraphics[width=\linewidth]{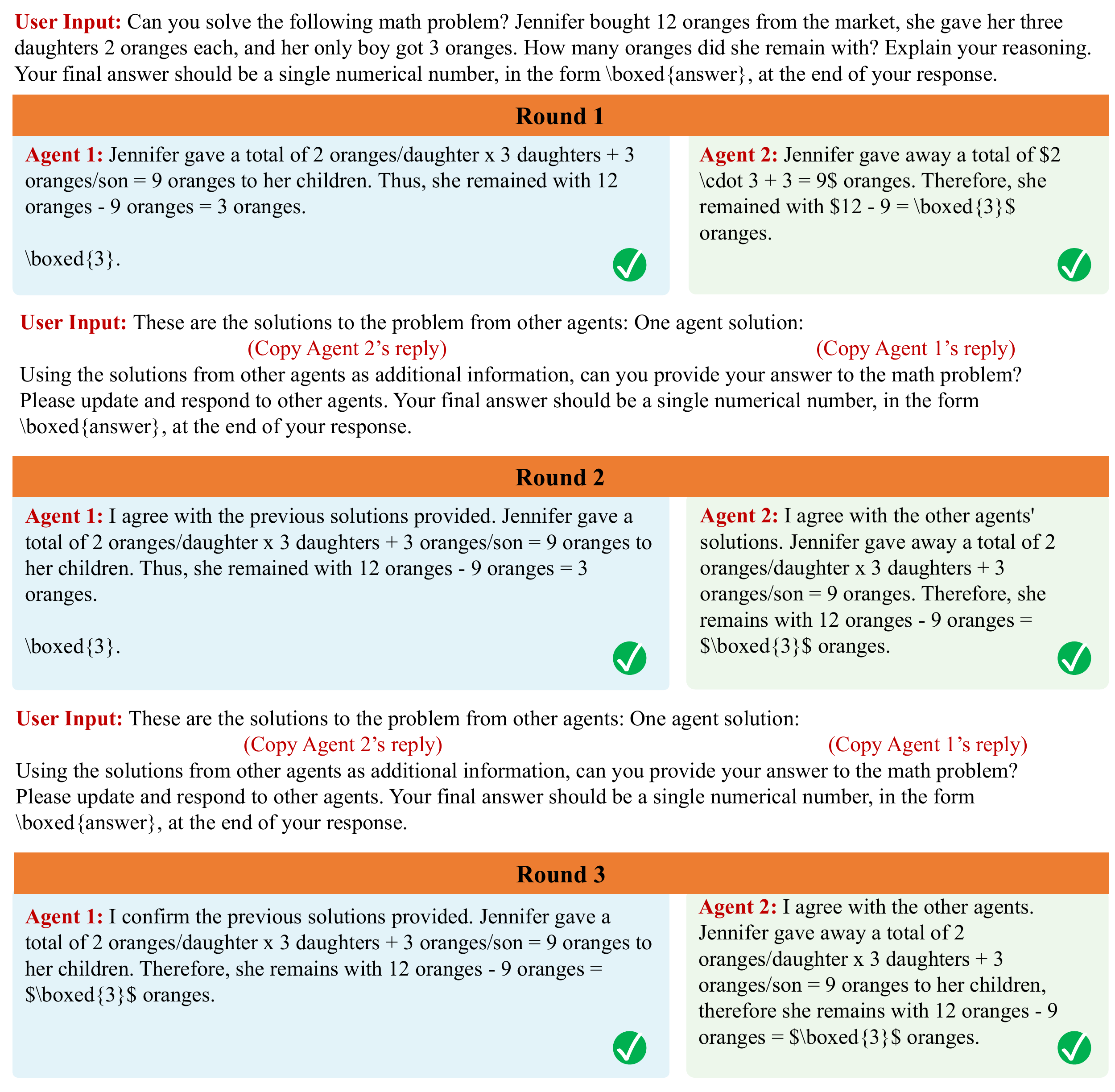}
    \caption{\textbf{Example of Correct GSM8K Debate.}}
    \label{fig:gsm-5}
    \vspace{-10pt}
\end{figure}

\begin{figure}[t]
    \centering
    \includegraphics[width=\linewidth]{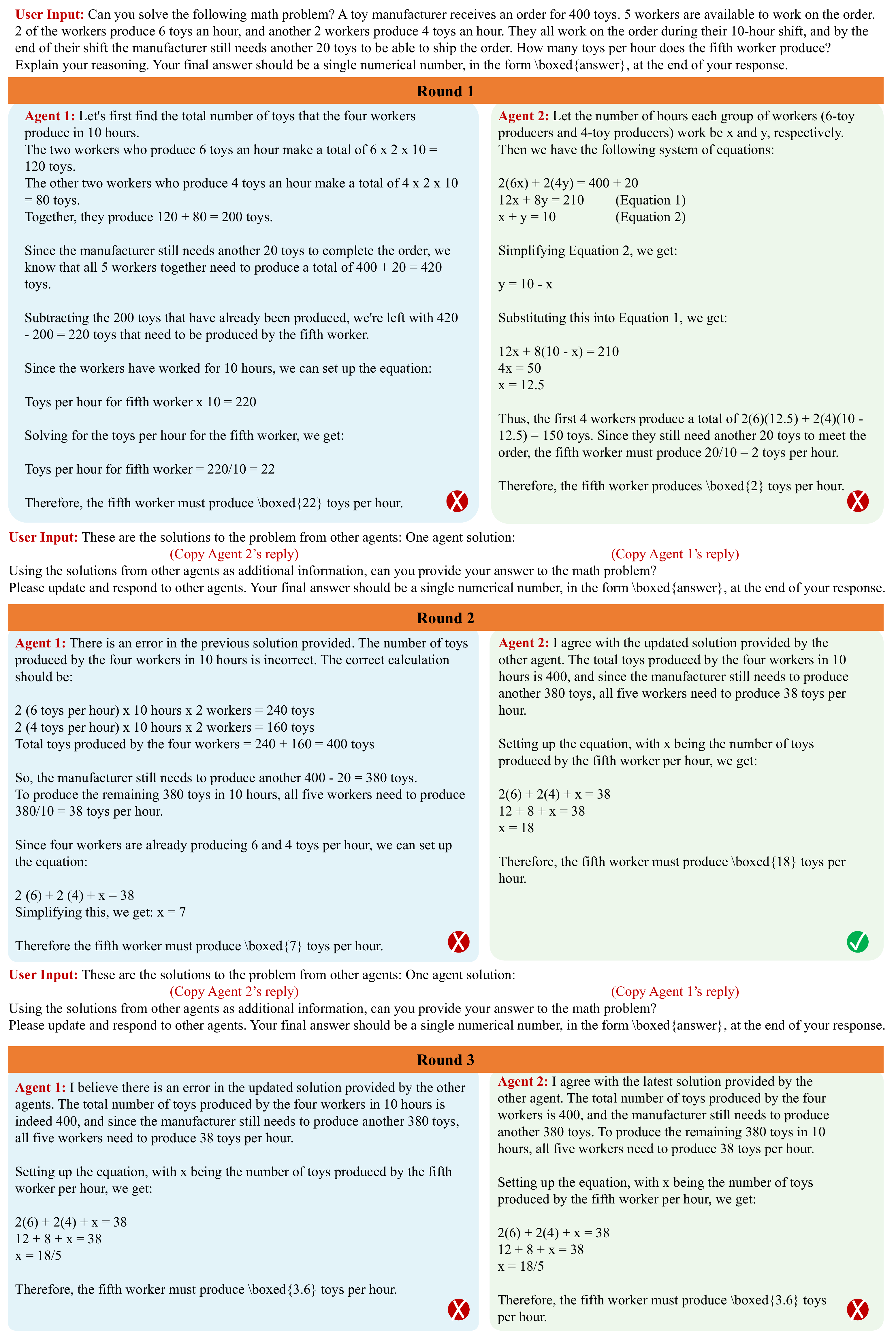}
    \caption{\textbf{Example of Incorrect GSM8K Debate.}}
    \label{fig:gsm-6}
    \vspace{-10pt}
\end{figure}

\begin{figure}[t]
    \centering
    \includegraphics[width=0.95\linewidth]{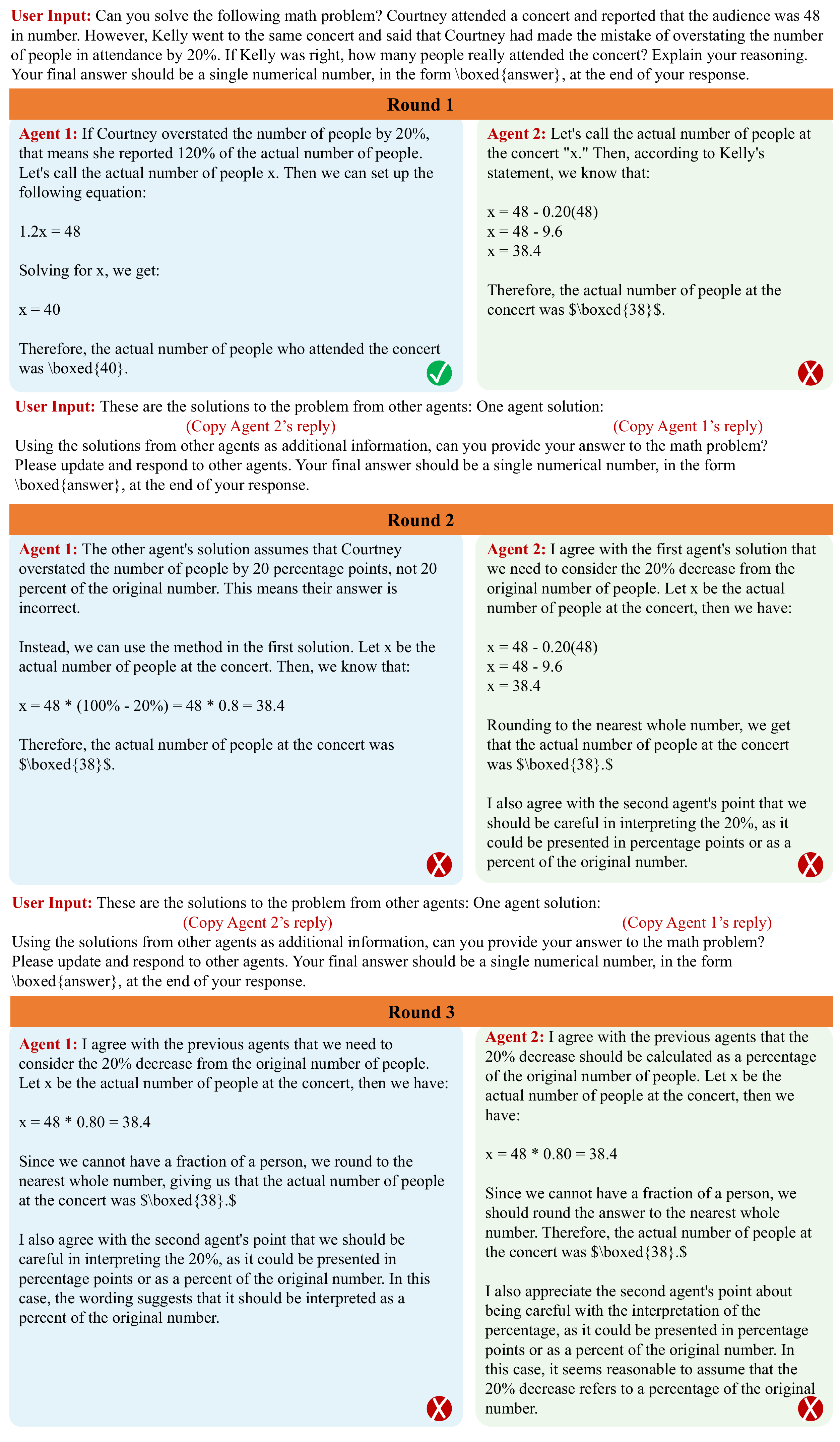}
    \caption{\textbf{Example of Incorrect GSM8K Debate.}}
    \label{fig:gsm-7}
    \vspace{-10pt}
\end{figure}

\begin{figure}[t]
    \centering
    \includegraphics[width=\linewidth]{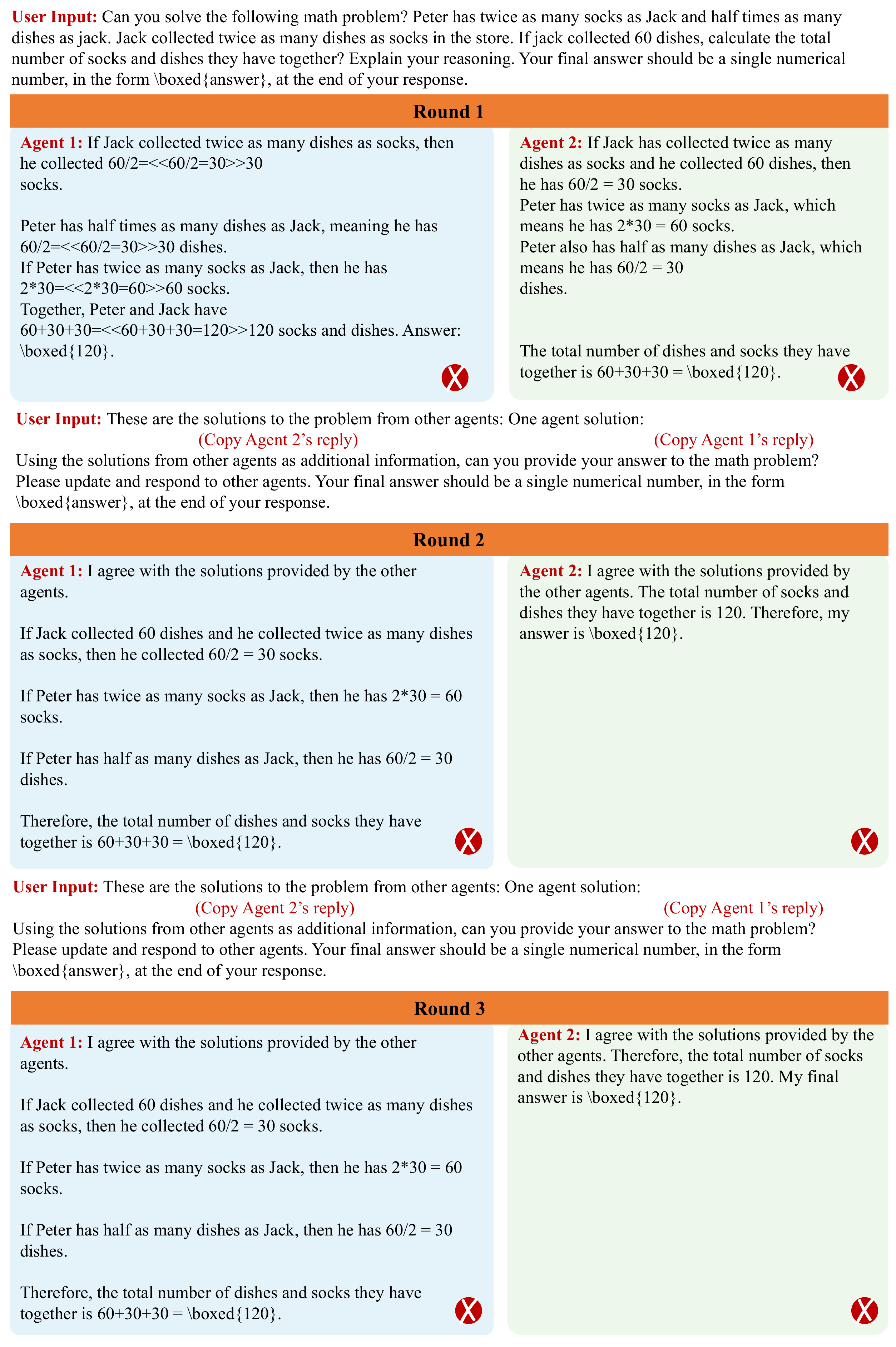}
    \caption{\textbf{Example of Incorrect GSM8K Debate.}}
    \label{fig:gsm-8}
    \vspace{-10pt}
\end{figure}

\begin{figure}[t]
    \centering
    \includegraphics[width=\linewidth]{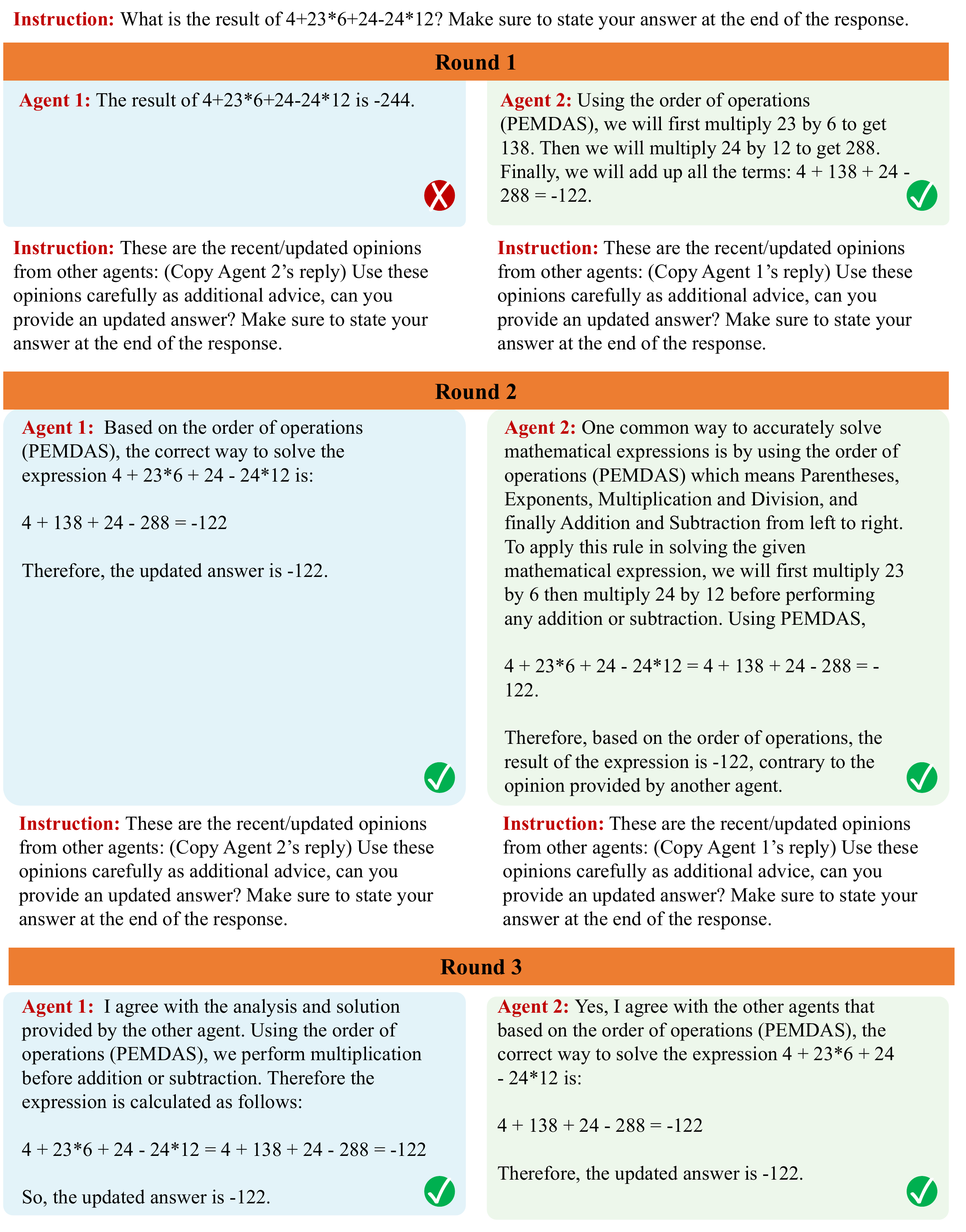}
    \caption{\textbf{Example of Arithmetic Debate.}}
    \label{fig:result_math}
    \vspace{-10pt}
\end{figure}

\begin{figure}[t]
    \centering
    \includegraphics[width=\linewidth]{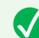}
    \caption{\textbf{Example of Arithmetic Debate with Summarization.} Four separate agents participate in debate, with two illustrated above. Instruction contains the summarized responses across agents.}
    \label{fig:result_math_summarize}
    \vspace{-10pt}
\end{figure}

\begin{figure}[t]
    \centering
    \includegraphics[width=\linewidth]{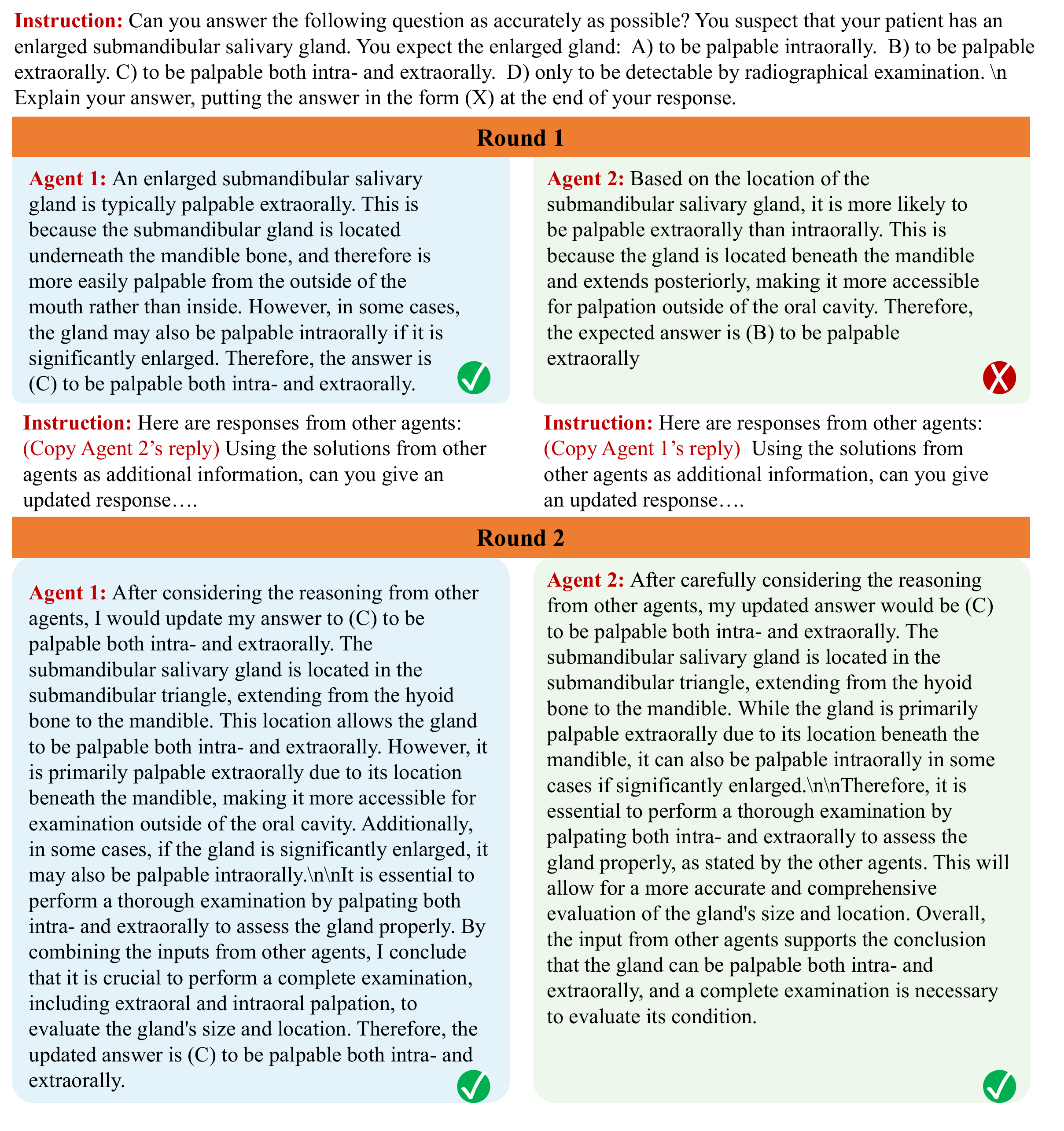}
    \caption{\textbf{Example of MMLU Debate.}}
    \label{fig:result_mmlu}
    \vspace{-10pt}
\end{figure}

\begin{figure}[t]
    \centering
    \includegraphics[width=\linewidth]{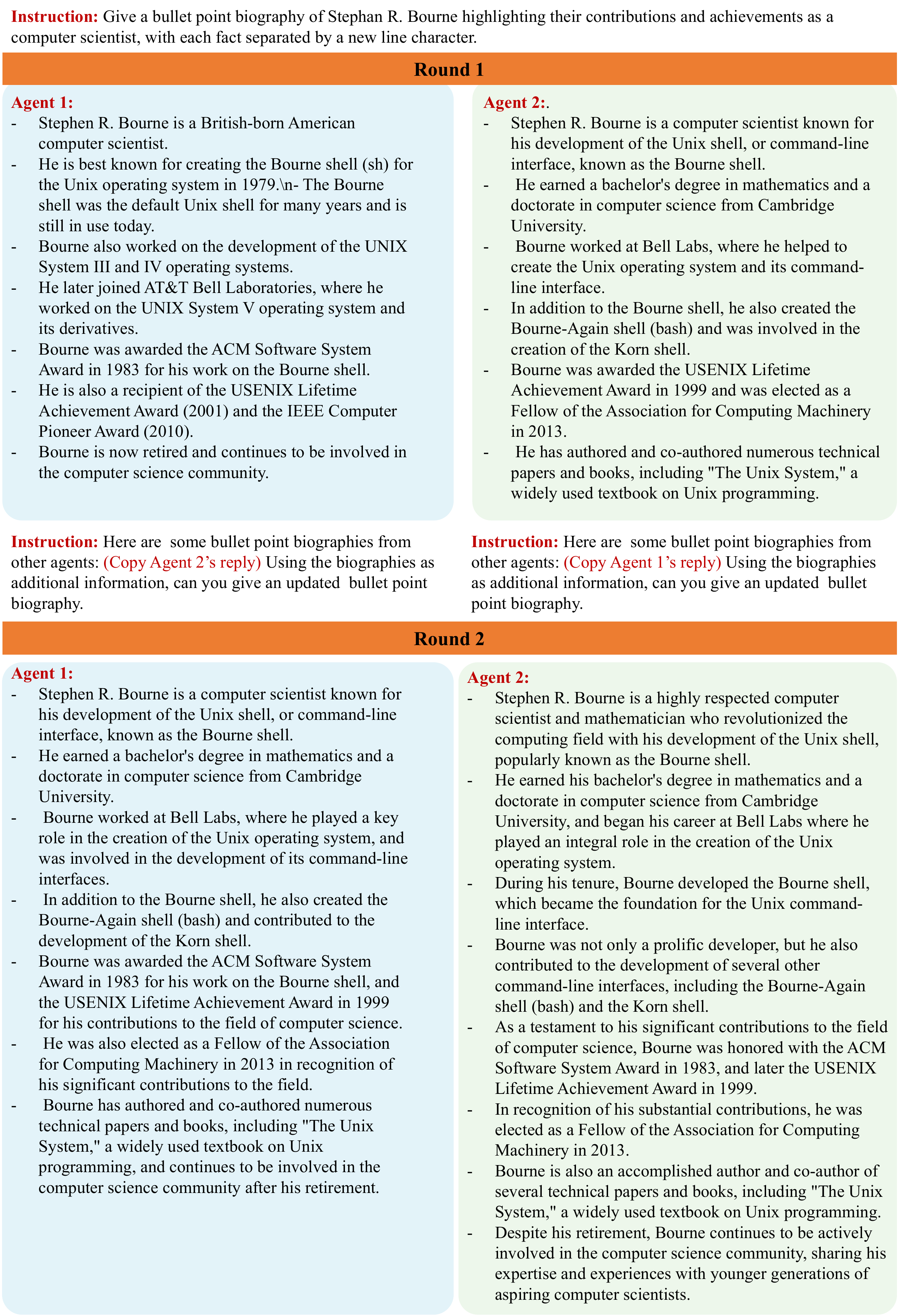}
    \caption{\textbf{Example of Biography Debate.} While we found that generated biographies after debate to be more accurate, many facts remain incorrect.}
    \label{fig:result_biography}
    \vspace{-10pt}
\end{figure}

\begin{figure}[t]
    \centering
    \includegraphics[width=\linewidth]{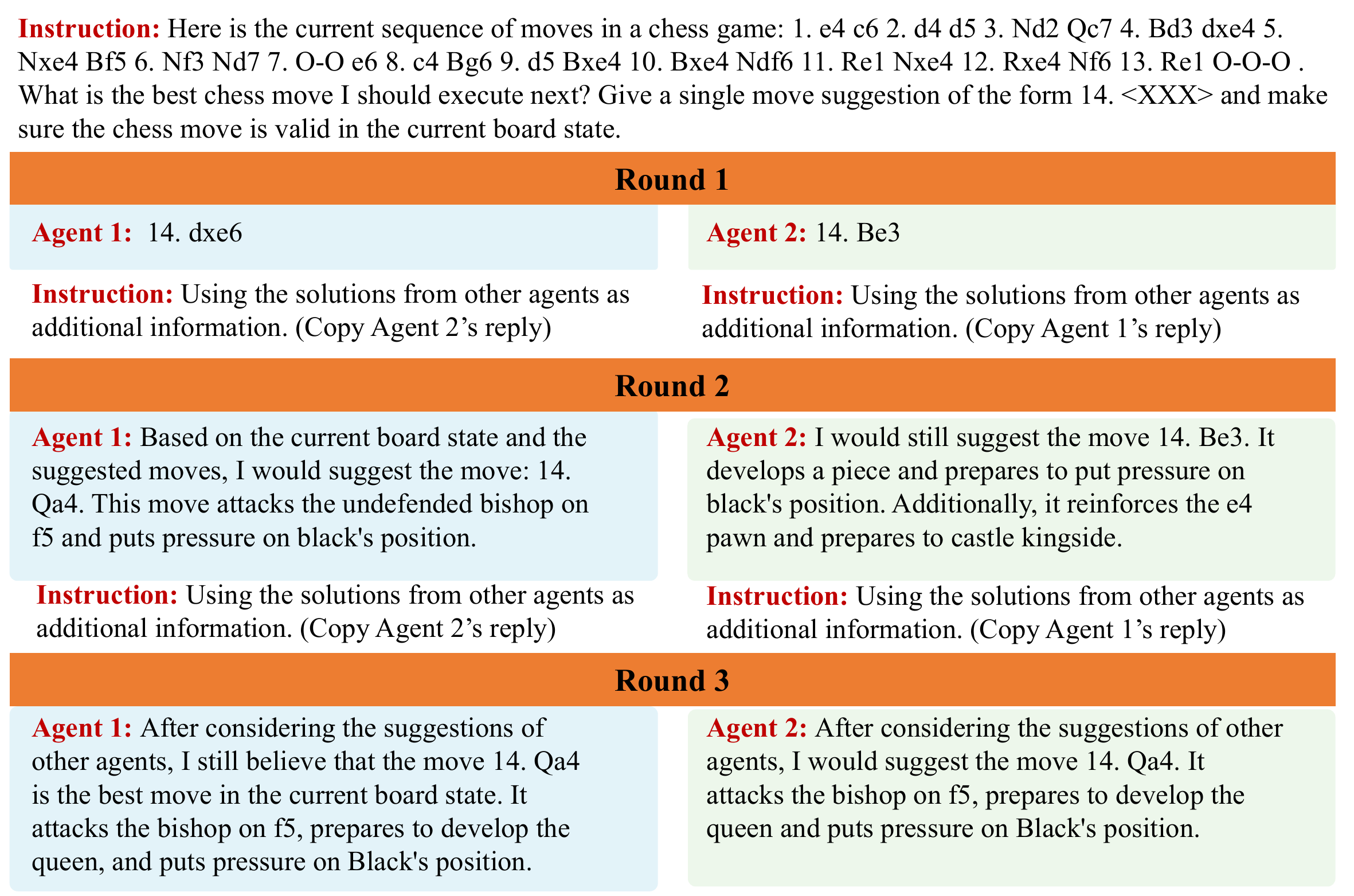}
    \caption{\textbf{Example of Chess Debate.}}
    \label{fig:result_chess}
    \vspace{-10pt}
\end{figure}

\myparagraph{Additional Qualitative Visualizations.} We added additional qualitative visualizations of the debate process. In \fig{fig:gsm-1}, \fig{fig:gsm-2}, \fig{fig:gsm-3}, \fig{fig:gsm-4}, \fig{fig:gsm-5}, we illustrate debates between agents in the GSM8K dataset which result in the correct answer. In \fig{fig:gsm-6}, \fig{fig:gsm-7}, \fig{fig:gsm-8}, we further illustrate debates in GSM8K which lead to the incorrect answer. We further provide an example illustration of debate in arithmetic in \fig{fig:result_math}, arithmetic with summarization of individual responses of agents in \fig{fig:result_math_summarize}, MMLU in \fig{fig:result_mmlu}, a debate with the full contents biographies in \fig{fig:result_biography}, and debate in chess in \fig{fig:result_chess}. In general, we found that debate improved the performance of final generated answers, though sometimes answers would converge to the incorrect value.

\subsection{Evaluation Details}
\label{sect:evaluation}

We provided detailed evaluation details for each setting in the paper. We run all experiments using the \texttt{gpt-3.5-turbo-0301} model. We provide a table listing the prompts used to prompt models and initialize debate in \tbl{tbl:prompt_appendix}.

\myparagraph{Arithmetic.} To evaluate the arithmetic task, we generated six random integers for each task between 0 and 30. We then evaluated the extent to which the correct integer answer was correctly obtained. We evaluated models on one hundred generated arithmetic tasks.

\myparagraph{Grade School Math.} To evaluate the GSM8K task, we evaluated the accuracy at which models were able to obtain the final correct answer, as extracted from a box. We evaluated models on one hundred grade school math problems.

\myparagraph{Chess.} To evaluate the chess reasoning task, we used chess games from \url{https://www.pgnmentor.com/players/Adams.zip}. We asked chatGPT to predict the next move for white to move at turn 14 and  reported the relative Stockfish pawn score with search depth 20 after executing the suggested move from chatGPT.  We evaluated models on three hundred selected chess games.

\myparagraph{Biographies.} To evaluate the biographies task, we compare each generated bullet point biography for a person with a ground truth set of facts about the person extracted from Wikipedia. We iteratively loop through each ground truth fact, and validate the extent to which the generated biography matches a particular bullet by prompting chatGPT with the prompt: \emph{Consider the following biography of <person>: <generated biography>  Is the above biography above consistent with the fact below? <ground truth bullet> Give a single-word answer, yes, no, or uncertain.} We then evaluate and report the percentage of ground bullets that chatGPT returns either yes or no on. We ignored ground truth bullets that chatGPT returns returned uncertain.

We found this evaluation metric provided a fast way to evaluate how relatively correct a generated bullet point biography is. However, we found that generated facts could contain incorrect information that was not captured in the ground truth bullet and thus could not be validated through this metric. Nevertheless, we believe this evaluation scheme estimates the relative accuracy of a generated biography.

\myparagraph{MMLU.} To evaluate MMLU, we measured the accuracy in which models were able to select the correct multiple-choice answer in each problem. We evaluated models on one hundred selected MMLU questions randomly distributed across each of the subject areas.

\myparagraph{Chess Validity.} To evaluate chess validity, we consider the BIG-Bench Chess-State Tracking Benchmark ~\cite{srivastava2022beyond}, where we used the hardest reported task in the benchmark \texttt{synthetic\_short}. Each generated answer was deemed correct as long as it was one of the valid answers in the sequence. We evaluated models of one hundred selected chess validity tasks.

\begin{table*}[t]
\small\setlength{\tabcolsep}{5.5pt}
\centering
\scalebox{0.8}{

\begin{tabular}{c|c|l}
     {\bf Task} & Type & {\bf Prompt} \\
      \midrule
     \multirow{4}{*}{Arithmetic}  & Starting & \emph{What is the result of \{\}+\{\}*\{\}+\{\}-\{\}*\{\}? Make sure to state your answer at the end of the response.} \\
     \cmidrule{2-3}
      & \multirow{3}{*}{Debate} & \emph{These are the recent/updated opinions from other agents: <other agent responses>  Use these opinions} \\
      & & \emph{carefully as additional advice, can you provide an updated answer? Make sure to state your answer} \\
      & & \emph{at the end of the response.} \\
      \midrule
      \multirow{6}{*}{GSM8K}  & \multirow{2}{*}{Starting} & \emph{Can you solve the following math problem? <Problem> Explain your reasoning. Your final answer }  \\
      & & \emph{should be a single numerical number, in the form \textbackslash boxed\{\{answer\}\}, at the end of your response.} \\
      \cmidrule{2-3}
     & \multirow{4}{*}{Debate} & \emph{These are the solutions to the problem from other agents: <other agent responses> Using the solutions} \\
     & & \emph{from other agents as additional information, can you provide your answer to the math problem? The original } \\
     & & \emph{math problem is <Problem>. Your final answer should be a single numerical number, in the form} \\
     & & \emph{ \textbackslash boxed\{\{answer\}\}, at the end of your response.} \\
      \midrule
      \multirow{2}{*}{Chess} & \multirow{3}{*}{Starting} & \emph{Here is the current sequence of moves in a chess game: <moves>. What is the best chess move I should  } \\
       &  & \emph{ execute next? Give a single move suggestion of the form 14. <XXX> and make sure the chess move } \\
       & & \emph{is valid in the current board state.} \\
     \cmidrule{2-3}
       & \multirow{3}{*}{Debate} & \emph{Here are other chess move suggestions from other agents: <other agent responses> Using the chess suggestions} \\
       & & \emph{ from other agents as additional advice and your earlier generated solution,  can you give me your updated thoughts  } \\
       & & \emph{ on the best next chess move I should play given the chess sequence {}? Give a single move suggestion of the form} \\
       & & \emph{ 14. <XXX> and make sure the chess move is valid in the current board state.} \\
      \midrule
      \multirow{4}{*}{Biographies} & \multirow{2}{*}{Starting} & \emph{Give a bullet point biography of {} highlighting their contributions and achievements as a computer scientist,} \\
      & & \emph{with each fact separated with a new line character.} \\
     \cmidrule{2-3}
       & \multirow{2}{*}{Debate} & \emph{Here are some bullet point biographies of <person> given by other agents: <other agent response> Closely} \\
       & & \emph{ examine your biography and the biography of other agents and provide an updated bullet point biography.} \\
      \midrule
      \multirow{5}{*}{MMLU} & \multirow{2}{*}{Starting} & \emph{Can you answer the following question as accurately as possible? {}: A) {}, B) {}, C) {}, D) {} Explain your answer, } \\
      & & \emph{putting the answer in the form (X) at the end of your response.} \\
     \cmidrule{2-3}
       & \multirow{3}{*}{Debate} & \emph{These are the solutions to the problem from other agents: <other agent responses> Using the reasoning}\\
       & & \emph{ from other agents as additional advice, can you give an updated answer? Examine your solution and}\\
       & & \emph{ that other agents. Put your answer in the form (X) at the end of your response.} \\
      \midrule
      \multirow{6}{*}{Chess Validity} & \multirow{3}{*}{Starting} & \emph{Given the chess game {}, give one valid destination square for the chess piece at {}. State the destination square } \\
      & & \emph{in the form (X), where X follows the regex [a-h][1-8], for example (e5). Give a one line explanation } \\
      & & \emph{of why your destination square is a valid move.} \\
     \cmidrule{2-3}
       & \multirow{3}{*}{Debate} & \emph{Here are destination square suggestions from other agents: <other agent responses>  Can you double}\\
       & & \emph{check that your destination square is a valid move? Check the valid move justifications from other agents.}\\
       & & \emph{State your final answer in a newline with a 2 letter response following the regex [a-h][1-8].} \\
    \bottomrule
\end{tabular}
}
\captionof{figure}{\small {\bf Prompts in each task.} List of prompts used in each task }
\label{tbl:prompt_appendix}
\vspace{-10pt}
\end{table*}

\end{document}